%% file: main.tex
\documentclass[lettersize,journal]{IEEEtran}
\usepackage{amsmath,amsfonts}
\usepackage{algorithmic}
\usepackage{algorithm}
\usepackage{array}
\usepackage[caption=false,font=normalsize,labelfont=sf,textfont=sf]{subfig}
\usepackage{textcomp}
\usepackage{stfloats}
\usepackage{url}
\usepackage{verbatim}
\usepackage{graphicx}
\usepackage{booktabs}
\usepackage{tabularx}

\usepackage[nocompress]{cite}
\usepackage{xcolor}   
\definecolor{mycite}{RGB}{35, 190, 192}
\definecolor{myfig}{RGB}{120,50,160}
\definecolor{mytab}{RGB}{0,120,90} 
\usepackage[
  colorlinks=true,
  citecolor=mycite,
  linkcolor=black,   
  urlcolor=blue,
  pdfborder={0 0 0}
]{hyperref}
\usepackage{cleveref}
\crefformat{figure}{\textcolor{myfig}{Fig.~#2#1#3}}
\crefformat{table}{\textcolor{mytab}{Tab.~#2#1#3}}

\usepackage{booktabs}
\usepackage{multirow}
\usepackage{enumitem}
\def\ie{\textit{i.e.}}
\def\eg{\textit{e.g.}}
\def\etal{\textit{et al.}}
 
\usepackage{pifont}
\usepackage{balance}

\usepackage{orcidlink}

\usepackage{colortbl}
\usepackage{makecell} 
\usepackage{wasysym}

\definecolor{highlightgray}{rgb}{0.95,0.95,0.95}

\usepackage{stfloats} 
\usepackage{caption} 

\usepackage{varwidth} 

\definecolor{remarkbg}{rgb}{0.927,1,1}
\definecolor{remarkborder}{gray}{0.15}
\colorlet{remarktitlebg}{remarkbg!20!black}
\usepackage[most]{tcolorbox}
\newenvironment{remark}[1][]{%
  \begin{tcolorbox}[
    enhanced,
    breakable,
    colback=remarkbg,          
    colframe=remarkborder,     
    boxrule=1pt,            
    arc=4pt,
    left=6pt,
    right=6pt,
    bottom=6pt,
    top=4pt,                  
    title={#1},
    fonttitle=\bfseries,
    coltitle=white,
    varwidth boxed title*=-2mm,
    boxed title style={
      colback=remarktitlebg,   
      colframe=remarkborder,
      boxrule=0.75pt,
      arc=2pt
    },
    attach boxed title to top left={
      xshift=6pt,
      yshift*=-\tcboxedtitleheight/2
    }
  ]
  \small\itshape
}{%
  \end{tcolorbox}
}
 
\newcommand{\nonumberfootnote}[1]{%
\begingroup
\renewcommand{\thefootnote}{}%
\footnote{#1}%
\endgroup
}

\begin{document}

\title{Image-to-Video Diffusion: From Foundations to Open Frontiers}
\pagestyle{plain}
 
\author{Xianlong Wang\orcidlink{0009-0009-3057-827X}, Wenbo Pan, Shijia Zhou, Ke Li, Yuqi Wang, Zeyu Ye, 
Hangtao Zhang,  Leo Yu Zhang\orcidlink{0000-0001-9330-2662}, and Xiaohua Jia\orcidlink{0000-0001-8702-8302},~\IEEEmembership{~IEEE Fellow}
\thanks{This paper was produced by the IEEE Publication Technology Group. They are in Piscataway, NJ.}
\thanks{Manuscript received April 19, 2021; revised August 16, 2021.}}

\markboth{Pre-print}%
{Image-to-Video Diffusion: From Foundations to Open Frontiers}


\twocolumn[{
    \renewcommand\twocolumn[1][]{#1}   
    \maketitle                       
    \centering                       
    \vspace{-0.8    cm}                  
    \includegraphics[width=0.9\linewidth]{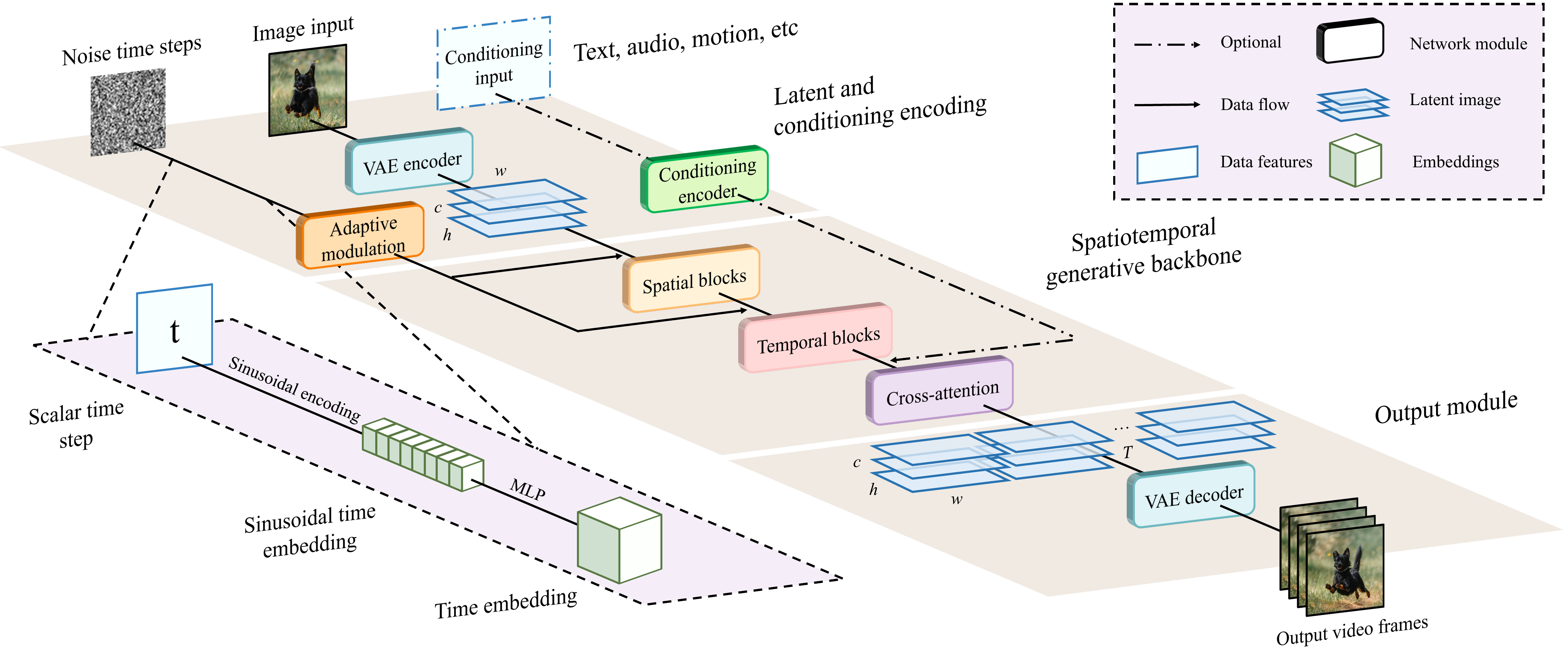}
    \captionsetup{type=figure, width=\linewidth}  
    \vspace{-0.22cm}          
    \captionof{figure}{Basic architecture of \textit{image-to-video} (I2V) diffusion models. }
    \label{fig:I2V-main}
    \vspace{0.3cm}                     
}]

\begin{abstract}
Diffusion-based \textit{image-to-video} (I2V) generation has become a central direction in generative models by turning a reference image, with optional conditions, into a temporally coherent video. 
Compared with broader video generation settings, this task places stricter demands on content consistency, identity preservation, and motion coherence. 
Although the literature grows rapidly, existing works mostly discuss I2V generation within broader topics and still lack a dedicated taxonomy together with a systematic analysis centered on this field. 
This work addresses that gap by treating diffusion I2V generation as a standalone subject. 
It first reviews the task formulation, model architectures, datasets, and evaluation metrics, and then organizes existing methods through a taxonomy based on architecture and training paradigm. 
It further distills four core designs, namely condition encoding, temporal modeling, noise prior design, and spatial-temporal upsampling, and discusses representative application scenarios together with major open challenges.  \nonumberfootnote{Xianlong Wang, Wenbo Pan, and Xiaohua Jia are with the Department of Computer Science, City University of Hong Kong, Hong Kong SAR, China  (e-mail: \{\href{mailto:xianlong.wang@my.cityu.edu.hk}{xianlong.wang},~\href{mailto:wenbo.pan@my.cityu.edu.hk}{wenbo.pan}\}\href{mailto:xianlong.wang@my.cityu.edu.hk}{@my.cityu.edu.hk};~\href{mailto:csjia@cityu.edu.hk }{csjia@cityu.edu.hk}).

Shijia Zhou and Hangtao Zhang are with the School of Cyber Science and Engineering, Huazhong University of Science and Technology, Wuhan 430074, China (e-mail:
~\href{mailto:hangt_zhang@hust.edu.cn}{hangt\_zhang@hust.edu.cn};~\href{mailto:shijiaz@hust.edu.cn}{shijiaz@hust.edu.cn}).

Ke Li is with the School of Cyber Science and Technology, University of Science and Technology of China, Hefei 230026, China (e-mail:~\href{mailto:like256@mail.ustc.edu.cn}{like256@mail.ustc.edu.cn}).

Yuqi Wang is with the Faculty of Science and Technology, Beijing Normal University-Hong Kong Baptist University, Zhuhai 519087, China (e-mail:~\href{mailto:u430036136@mail.bnbu.edu.cn}{u430036136@mail.bnbu.edu.cn}).

Zeyu Ye is with the School of Computer Science, Xiangtan University, XiangTan 411105, China (e-mail:~\href{mailto:202305566820@smail.xtu.edu.cn}{202305566820@smail.xtu.edu.cn}). 

Leo Yu Zhang is with the School of Information and Communication
Technology,  Griffith University, Southport, QLD 4215, Australia (e-mail: \href{mailto:leo.zhang@griffith.edu.au}{leo.zhang@griffith.edu.au}). } 
\end{abstract}

\setcounter{footnote}{0}

\begin{IEEEkeywords}
Image-to-video, video generation, diffusion models, generative models, deep learning.
\end{IEEEkeywords}

\input{sections/sec1_introduction}

\section{Foundations of Diffusion I2V}
\label{sec:preliminaries}

\subsection{Diffusion I2V Formulation}
\label{sec:i2v-definition}

\subsubsection{Definition} Diffusion-based 
\textit{image-to-video } (I2V)~\cite{blattmann2023stable,singer2023make,niu2024mofa} generation aims to synthesize a temporally coherent video clip $V_{\text{gen}}$ that is faithful to a given reference image while exhibiting plausible motion dynamics.
Formally, given a reference image $I_{\text{in}} \in \mathbb{R}^{H \times W \times 3}$, an optional text prompt $\tau$~\cite{lin2025mvportrait}, and a set of optional control signals $\mathbf{c}$~\cite{li2025realcam}, an I2V model learns a conditional distribution as follows:
\begin{equation}
p_{\theta}(V_{\text{gen}} \mid I_{\text{in}}, \tau, \mathbf{c}), 
\quad V_{\text{gen}} = \{v_k\}_{k=1}^{K},\ v_k \in \mathbb{R}^{H \times W \times C}
\end{equation}
where $v_k$ denotes the $k$-th RGB frame of the generated video with spatial resolution $H \times W$.  
Most modern I2V systems~\cite{liu2023pilife,lin2025mvportrait,lin2025stiv,guo2024i4vgen} instantiate $p_{\theta}$ via conditional diffusion models that iteratively refine a noisy video toward a clean sample, while preserving appearance induced by $I_{\text{in}}$ and $\tau$.

\subsubsection{Video Representation}
To improve computational efficiency, existing I2V models~\cite{blattmann2023stable,ni2023conditional,zhang2025tora,liu2024mardini,hacohen2024ltx} commonly perform diffusion in a compact \emph{latent video space} using an autoencoder (\eg, VAE~\cite{cheng2025leanvae}) that maps pixels to latents and back. 
Specifically, an encoder $\mathcal{E}(\cdot)$ compresses a pixel-space video sample $\mathbf{V}_0$  into latent variables $\mathbf{Z}_0=\mathcal{E}(\mathbf{V}_0)$, and the diffusion process is performed in the latent space  rather than in pixel space. After denoising, a decoder $\mathcal{D}(\cdot)$ maps the recovered latents back to the video domain, yielding $\mathbf{V}_0=\mathcal{D}(\hat{\mathbf{Z}}_0)$. 
Accordingly, the denoiser is trained with a latent noise-prediction objective, formulated as:
\begin{equation}
\mathcal{L}_{\mathrm{i2v}}(\theta)=
\mathbb{E}_{\mathbf{Z}_0,\, t,\, \boldsymbol{\epsilon}}
\left[
\left\|
\boldsymbol{\epsilon}-
\boldsymbol{\epsilon}_{\theta}(\mathbf{Z}_t,t;I_{\text{in}},\tau,\mathbf{c})
\right\|_2^2
\right]
\end{equation}
where $\mathbf{Z}_t=\sqrt{\bar{\alpha}_t}\mathbf{Z}_0+\sqrt{1-\bar{\alpha}_t}\boldsymbol{\epsilon}$ follows the forward noising process in latent space. 
In practice, the conditioning signals $(I_{\text{in}},\tau,\mathbf{c})$ are fused into the denoiser via cross-attention~\cite{lin2025stiv,shi2024motioni2v} or feature injection~\cite{xing2024dynamicrafter}, enabling the model to preserve the reference image while synthesizing temporal consistency.

\subsubsection{Conditioning Mechanisms}
Beyond basic I2V generation~\cite{guo2024i4vgen,shi2024motioni2v,liu2023pilife,tian2025reducio,wan2025open}, practical schemes~\cite{xu2024camco,shi2024motioni2v,tian2024emo,low2025talkingmachines} often expose additional control via $\mathbf{c}$, as shown in~\cref{fig:image-condition}.  
The following are the representative I2V conditions: 
\begin{itemize}[label=\ding{117}]
\item \textbf{Image-only:} 
Early I2V methods establish the foundation by conditioning solely on reference images. 
SVD~\cite{blattmann2023stable} exemplifies image-only conditioning by learning temporal dynamics without textual guidance. Building on this, I2VGen-XL~\cite{zhang2023i2vgenxl} adopts a cascaded design that decouples semantic understanding from detail refinement, while LFDM~\cite{ni2023conditional} introduces latent flow diffusion to model motion-content separation. 
Additional schemes exploring pure image conditioning include TRIP~\cite{zhang2024trip}, AtomoVideo~\cite{gong2024atomovideo}, and the training-free scheme TI2V-Zero~\cite{ni2024ti2v}. 
However, image-only conditioning faces a  limitation, \ie, models cannot interpret user intent regarding desired scenes, restricting practical applicability. 

\item \textbf{Image+Text:} 
Textual conditions enable semantic control over generated dynamics. 
Therefore, DynamiCrafter~\cite{xing2024dynamicrafter} pioneers this dual conditioning paradigm through a query transformer that projects images into text-aligned feature spaces, enabling the model to understand what motion would be semantically appropriate for a given scene.  
Following this, ConsistI2V~\cite{ren2024consisti2v} enhances visual consistency through spatiotemporal attention mechanisms, while Emu Video~\cite{girdhar2024emu} factorizes video generation into separate T2I and I2V stages. 
Moreover, VideoCrafter1~\cite{chen2023videocrafter1} and VideoCrafter2~\cite{chen2024videocrafter2} provide toolkits supporting both text and image conditions. 
For large models, Open-Sora~\cite{zheng2024opensora} and Open-Sora Plan~\cite{lin2024opensora} demonstrate that DiT can integrate text conditioning seamlessly across billions of parameters. 
Industrial-scale implementations further verify this route, with CogVideoX~\cite{yang2025cogvideox} employing expert adaptive layer normalization for text-video fusion, HunyuanVideo~\cite{kong2024hunyuanvideo} reaching 13 billion parameters through dual-stream processing, and Wan~\cite{wan2025open} and Seedance~\cite{gao2025seedance} pushing model scale even further.

\item \textbf{Image+Motion:} Motion provides an explicit motion specification beyond appearance (\eg, motion fields, optical flow), so I2V models can follow a dynamic pattern while keeping the image content. 
A first line of work represents motion as dense fields, where Motion-I2V~\cite{shi2024motioni2v} predicts a motion field and performs motion-guided temporal attention, and MOFA-Video~\cite{niu2024mofa} densifies sparse motion cues into motion fields. 
A second line focuses on human motion~\cite{zhang2025mimicmotion,gan2025humandit}, using pose sequences through dedicated guidance/adapters.  
Beyond flow/pose, motion can also be specified through geometric constraints, such as bounding-box trajectories~\cite{wang2024boximator} and reference transfer from a reference video for zero-shot motion imitation~\cite{chen2025lmp}. 
Finally, several works exploit auxiliary cues or model-level control, \ie, intermediate motion-related signals~\cite{liang2024movideo} and adjusting motion degree via model merging~\cite{tian2025extrapolating}, while CustomCrafter~\cite{wu2025customcrafter} and Loopy~\cite{jiang2025loopy} extend motion conditioning to personalization and audio-driven portrait animation, respectively.

\begin{figure}[t]
    \centering
    \includegraphics[width=1\linewidth]{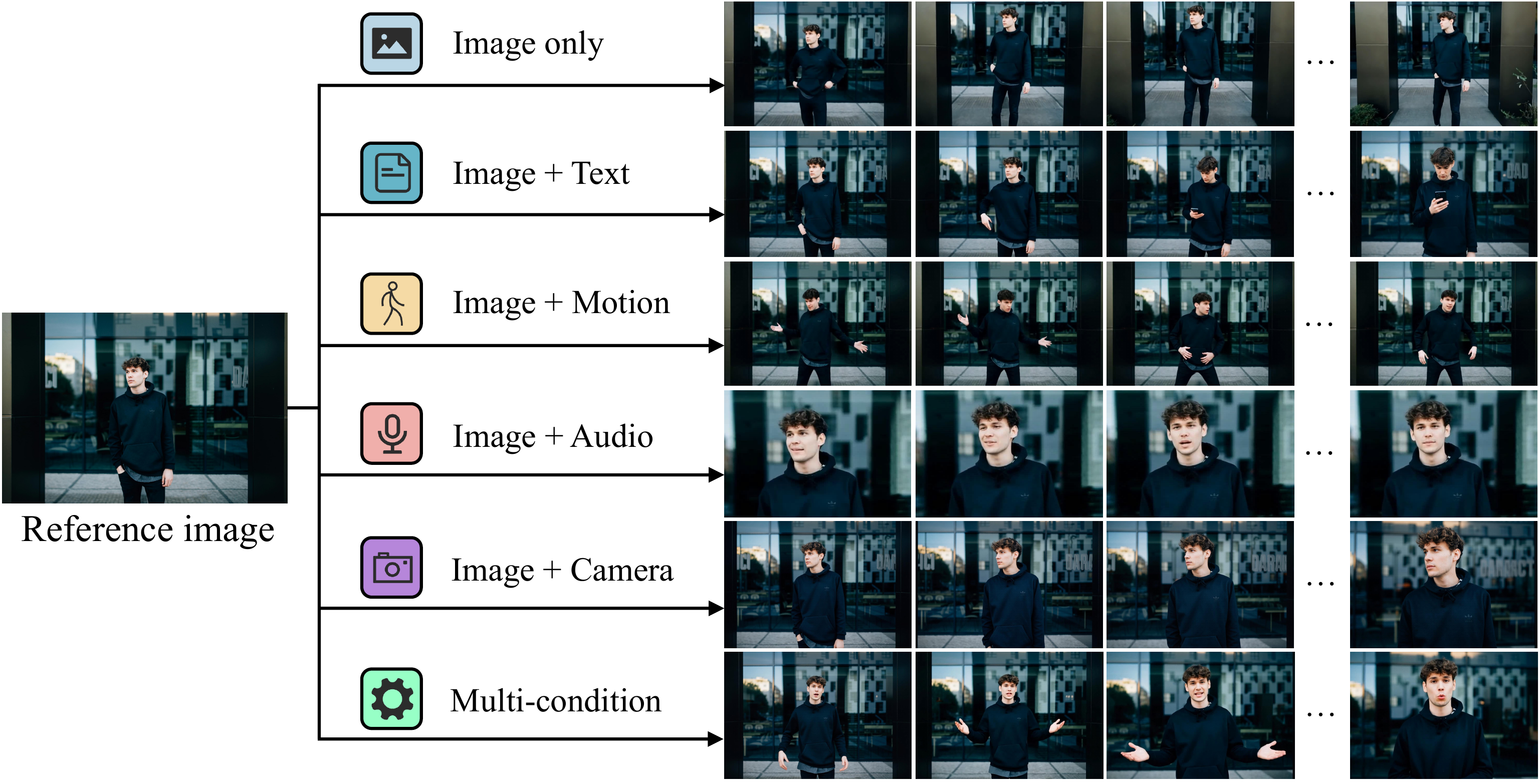}
    \caption{Video effects of I2V models under different conditions.}
    \label{fig:image-condition}
\end{figure}

\item \textbf{Image+Audio:} Audio conditioning treats speech as a time-varying control signal, thereby generating motion-consistent dynamics such as lip, facial expression, and head motion. 
EMO~\cite{tian2024emo} directly performs audio-to-video  under weak conditions. Afterwards, some works improve controllability by structuring audio-to-motion effects, such as decomposing audio cues at different semantic levels~\cite{xu2024hallo}, scaling to long-duration generation~\cite{cui2024hallo2}, and emphasizing global audio perception to capture prosody-driven dynamics~\cite{ji2025sonic}.  
Additionally, EchoMimic~\cite{chen2025echomimic} introduces editable facial landmarks that enable hybrid audio-landmark conditioning, while AniPortrait~\cite{wei2024aniportrait} targets photorealistic audio-driven portrait animation. 
For long-horizon and streaming settings, Loopy~\cite{jiang2025loopy} models long-term audio-guided motion, StableAvatar~\cite{tu2025stableavatar} extends audio conditioning to  unbounded avatar generation, and TalkingMachines~\cite{low2025talkingmachines} achieves audio-driven ``FaceTime-style'' video via autoregressive design, with Omniavatar~\cite{gan2025omniavatar} focusing on efficient and adaptive body animation under audio guidance. 
Finally, audio conditioning is extended from heads to bodies and multi-characters, where CyberHost~\cite{lin2025cyberhost} uses a one-stage diffusion for audio-driven talking-body generation, EMO2~\cite{tian2025emo2} augments audio control with end-effector guidance for stronger motion constraints, and Hunyuan Video-Avatar~\cite{chen2025hunyuanvideo} targets high-fidelity audio-driven human animation for characters.

\item \textbf{Image+Camera:} Camera conditioning in  I2V specifies an explicit camera trajectory as a control signal, so that generation follows the desired viewpoint. 
CamCo~\cite{xu2024camco} and CamI2V~\cite{zheng2024cami2v} encode pose/ray-based camera representations and inject them into the denoiser with geometry-aware  constraints. 
RealCam-I2V~\cite{li2025realcam} further emphasizes real-world controllability by reconstructing a metric 3D proxy from the input image.  From the data side, HumanVid~\cite{wang2024humanvid} provides camera-motion annotations to supervise camera-controllable animation. 
For static scenes, SRENDER~\cite{chen2026efficient} achieves efficient video generation via sparse trajectory-conditioned keyframes and 3D reconstruction.

\item \textbf{Image+Multi-condition:} 
As individual conditioning modalities mature, the requirement of multi-dimension control pushes the development of schemes capable of handling mixed conditions. 
OmniHuman-1~\cite{lin2025omnihuman} exemplifies this trend through mixed training on text, audio, video, and pose conditions with adaptive strength hierarchies, enabling flexible control over a combination of available modalities. 
MoonShot~\cite{zhang2024moonshot} provides a multimodal conditioning framework, while MOFA-Video~\cite{niu2024mofa} combines motion field adaptations from different sources and MegActor-Sigma~\cite{yang2025megactor} unlocks joint control across diverse inputs. 
Additionally,  physics-grounded generation represents an emerging direction where physical constraints guide synthesis. 
PhysGen~\cite{liu2024physgen} combines rigid-body physics simulation with diffusion models in a training-free manner, PhyRPR~\cite{zhao2026phyrpr} integrates physical reasoning into the generation process, and  
VACE~\cite{jiang2025vace} presents an all-in-one framework unifying multiple   tasks through a universal conditioning architecture. 

\end{itemize}

In summary, practical I2V diffusion models expose controls $\mathbf{c}$ beyond basic image references. 
Such $\mathbf{c}$ typically specifies camera trajectories, motion cues, or audio as a temporal driver. These signals are injected via conditioning or guidance modules to achieve viewpoint-consistent geometry, controllable dynamics, and speech-synchronized animation.

\input{sections/sec2_3_architectures}

\input{sections/sec2_4_datasets_metrics}

\section{Taxonomy of Image-to-Video Methods}
\label{sec:taxonomy}
\begin{figure*}[!h]
  \centering
  \includegraphics[width=1\linewidth]{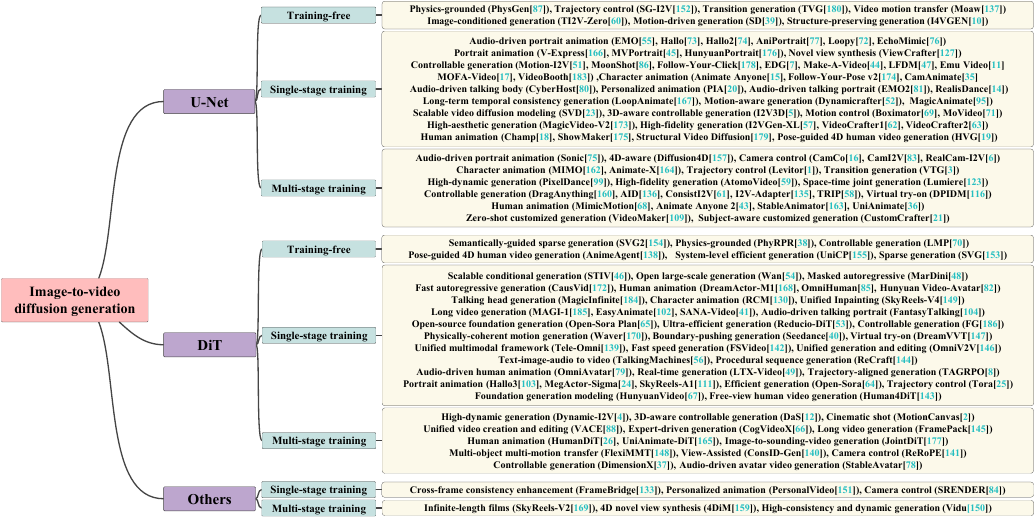}
  \caption{A taxonomy of diffusion I2V schemes, organized by model architecture and then subdivided by training paradigm. }
  \label{fig:I2V-taxmony}
\end{figure*}

\input{sections/sec3_1_condition_type}

\input{sections/sec3_2_training_paradigm}

\section{Key Design in Diffusion I2V}
\label{sec:core-components}

\input{sections/sec4_1_2_encoding_temporal}

\input{sections/sec4_3_5_noise_upsample_ctrl}

\section{Applications of I2V Models}
\label{sec:applications}

\input{sections/sec7_1_human_animation}

\input{sections/sec8_open_problems}

\input{sections/sec9_conclusion}

\balance
\bibliographystyle{IEEEtran}
\bibliography{ref}
\vspace{-3.5em}

\begin{IEEEbiographynophoto}{Xianlong Wang} is currently pursuing the Ph.D. degree with the Department of Computer Science, City University of Hong Kong, Hong Kong.
His research interests include video generation and diffusion models. 
His work has been published at venues like NeurIPS, ICLR, AAAI, and ACM MM. 
He served as a reviewer for CVPR, ICML, NeurIPS, ICLR, AAAI, and ACM MM. 
\end{IEEEbiographynophoto}
\vspace{-3.5em}

\begin{IEEEbiographynophoto}{Wenbo Pan} is currently pursuing the Ph.D. degree with the Department of Computer Science, City University of Hong Kong, Hong Kong. His research focuses on   computer vision and 
alignment of (multi-modal) large language  models. 
His work has been published at venues including ICML, ICLR, and EMNLP. 
He served as a reviewer for CVPR, ICML,  and ICLR.
\end{IEEEbiographynophoto}
\vspace{-3.5em}

\begin{IEEEbiographynophoto}{Shijia Zhou} is currently pursuing the B.S. degree with Huazhong University of
Science and Technology. 
Her research focuses on video generation. 
\end{IEEEbiographynophoto}
\vspace{-3.5em}

\begin{IEEEbiographynophoto}{Ke Li} is currently pursuing the B.S. degree through a joint training program of Southwest University of Science and Technology and University of Science and Technology of China. 
Her research focuses on video diffusion models.   
\end{IEEEbiographynophoto}
\vspace{-3.5em}

\begin{IEEEbiographynophoto}{Yuqi Wang} is  pursuing the B.S. degree with  Beijing Normal University-Hong Kong Baptist University. 
His research focuses on  video synthesis. 
\end{IEEEbiographynophoto}
\vspace{-3.5em}

\begin{IEEEbiographynophoto}{Zeyu Ye} is currently pursuing the B.S. degree with  Xiangtan University. 
His research focuses on   human image animation and diffusion models. 
\end{IEEEbiographynophoto}
\vspace{-3.5em}

\begin{IEEEbiographynophoto}{Hangtao Zhang} is currently pursuing the M.S. degree with  Huazhong University of
Science and Technology. His research interests include object detection, diffusion generation. 
His research has been published in CVPR, NeurIPS, and ICLR. He served as a reviewer for CVPR, ECCV, NeurIPS, ICML, AAAI,  and ICLR.
\end{IEEEbiographynophoto}
\vspace{-3.5em}

\begin{IEEEbiographynophoto}{Leo Yu Zhang} received the PhD degree from the City University of Hong Kong, Hong Kong, in 2016.
He has also held various research positions with the City University of Hong Kong, the University of Macau, the University of Ferrara, and the University of Bologna. 
From 2018 to 2023, he was a faculty member with the School of Information Technology,
Deakin University. He is currently an associate professor with the School of Information and Communication
Technology, Griffith University, QLD, Australia. 
He has authored or coauthored more than 100 articles in refereed journals and conferences, such as CVPR, ICCV, 
NeurIPS, ICML, ICLR, and AAAI. His research interests include trustworthy
AI and computer vision.
\end{IEEEbiographynophoto}
\vspace{-3.5em}

\begin{IEEEbiographynophoto}{Xiaohua Jia}
(IEEE Fellow, ACM Fellow) received the B.Sc. and M.E. degrees in 1984 and 1987, respectively, from
the University of Science and Technology of China,
and the Ph.D. degree in 1991 in Information Science from the University of Tokyo. He is currently the Chair Professor with the Department of Computer Science at City University of Hong Kong. 
His research interests include generative artificial intelligence, 
data privacy and security, cloud computing, 
and distributed networks. 
He was an editor of the IEEE TPDS, IEEE IoTJ, Wireless Networks, Journal of World Wide Web, Journal of Combinatorial Optimization. 
He was the general chair of ACM MobiHoc 2008, TPC co-chair of IEEE MASS
2009, area-chair of IEEE INFOCOM 2010, TPC co-chair of IEEE GlobeCom 2010-Ad Hoc and Sensor Networking Symposium, and Panel co-chair of IEEE INFOCOM 2011. He is fellow of the IEEE Computer Society, and
Fellow of the ACM. 
\end{IEEEbiographynophoto}

\end{document}

%% file: sections/sec1_introduction.tex

\section{Introduction}
\label{sec:introduction}

\IEEEPARstart{T}{he} rapid advancement of generative models has fundamentally transformed content creation, with diffusion-based \textit{image-to-video} (I2V) generation~\cite{liu2025dynamic,wang2025levitor,xing2025motioncanvas,yang2025vtg,zhang2025i2v3d,li2025realcam,tian2025extrapolating,wang2026tagrpo} emerging as a particularly compelling frontier. 
Given a static reference image and optional conditioning signals such as text prompts, motion trajectories, or audio, I2V models synthesize temporally coherent video clips that faithfully preserve the appearance of the input while exhibiting plausible dynamics.
As shown in~\cref{fig:I2V-main}, the I2V diffusion architecture primarily comprises a \textit{Variational Autoencoder} (VAE)~\cite{kingma2013auto} encoder for latent and conditioning encoding, spatial and temporal blocks for spatiotemporal processing, cross-attention mechanisms for feature integration, and a VAE decoder for output video frame generation. 
Compared with \textit{text-to-video} (T2V)~\cite{guo2024i4vgen,girdhar2024emu} or unconditional video generation~\cite{gupta2022rv,wang2023styleinv}, I2V places stricter constraints on content consistency, identity preservation, and temporal coherence, making it valuable for applications such as character animation~\cite{zhou2024realisdance,hu2024animate}, controllable viewpoint synthesis~\cite{xu2024camco,niu2024mofa,zhu2024champ,wang2026human}, and personalized generation~\cite{zhang2024pia,wu2025customcrafter}. 
As foundation models continue to scale, the generative backbone of  I2V diffusion models has progressed from early U-Net prototypes~\cite{ronneberger2015unet,blattmann2023stable} to \textit{Diffusion Transformer} (DiT)~\cite{yang2025megactor,zhang2025tora,gan2025humandit}, serving as a central topic in video generation.

Despite the explosive growth of diffusion I2V literature, existing works provide only partial coverage of this landscape.  
In particular, they mainly analyze diffusion models from broader perspectives, such as general video generation~\cite{xing2024survey,wang2025survey}, controllable generation~\cite{ma2025controllable}, video editing~\cite{sun2024diffusion}, efficient diffusion modeling~\cite{ma2025efficient}, T2V generation~\cite{kumar2025bridging}, human-centric generation~\cite{xue2025human}, or diffusion methods in general~\cite{yang2023diffusion}. 
While these works provide useful overviews of related tasks, architectures, and applications, I2V is usually discussed only as one component within a larger landscape. 
More importantly, even the most relevant prior works only partially cover key I2V dimensions, such as content consistency, motion modeling, training data, and evaluation metrics, and they rarely organize these aspects under an I2V-specific framework~\cite{xing2024survey,wang2025survey,ma2025controllable}. 
As a result, these gaps motivate a focused review that treats I2V diffusion generation as a first-class topic and provides a more complete taxonomy and systematic analysis of its technical foundations and future directions.

In response,  as seen in~\cref{fig:I2V-taxmony}  
we build a two-level taxonomy that organizes  first by model architecture, \eg, U-Net~\cite{wang2024humanvid,wang2025unianimate}, DiT~\cite{sun2024dimensionx,wang2026tagrpo}, and then by training paradigm, \eg, training-free~\cite{zhao2026phyrpr,liu2023pilife}, multi-stage~\cite{gao2025seedance,chen2025sana}, and fine-tuning~\cite{zhang2025framepack,hu2025animate} schemes. 
Simultaneously, 
we  provide a detailed review of the definition, architectures, datasets, and evaluation metrics of I2V diffusion, offering a comprehensive understanding of this field. 
Furthermore, to study the foundation techniques in I2V, we analyze four key designs, \ie, condition encoding, temporal modeling, noise prior designs, and spatial-temporal upsampling, spanning the workflow of I2V diffusion modeling. 
We also discuss the practical application scenarios of I2V from three aspects: animation, viewpoint control, and media production. 
Finally, we discuss existing I2V schemes through four key open challenges including temporal consistency, video controllability, efficient deployment, and model security to propose promising future research directions.

To offer a preliminary glimpse, this research is organized as follows: 
Section~\ref{sec:preliminaries} outlines technical foundations of diffusion I2V, including formulation, architectures, datasets, and evaluation metrics, providing a holistic foundation for understanding. 
Next, 
Section~\ref{sec:taxonomy} presents a comprehensive taxonomy of I2V generation schemes, categorized by architecture and training paradigm. Section~\ref{sec:core-components} 
provides a detailed discussion of the key designs within existing I2V technical architectures. 
In Section~\ref{sec:applications}, we explore the practical deployment of I2V models in real-world settings.
In Section~\ref{sec:open_problems}, we discuss several open issues within the I2V domain and outline key trends for future. 
Finally, Section~\ref{sec:summary} makes a summary of this paper. 
Note that all
references in this paper are current as of March 20, 2026.

%% file: sections/sec2_3_architectures.tex
\subsection{Basic Architectures}
\label{sec:architectures}

\begin{figure*}[t]
    \centering
    \includegraphics[width=1\linewidth]{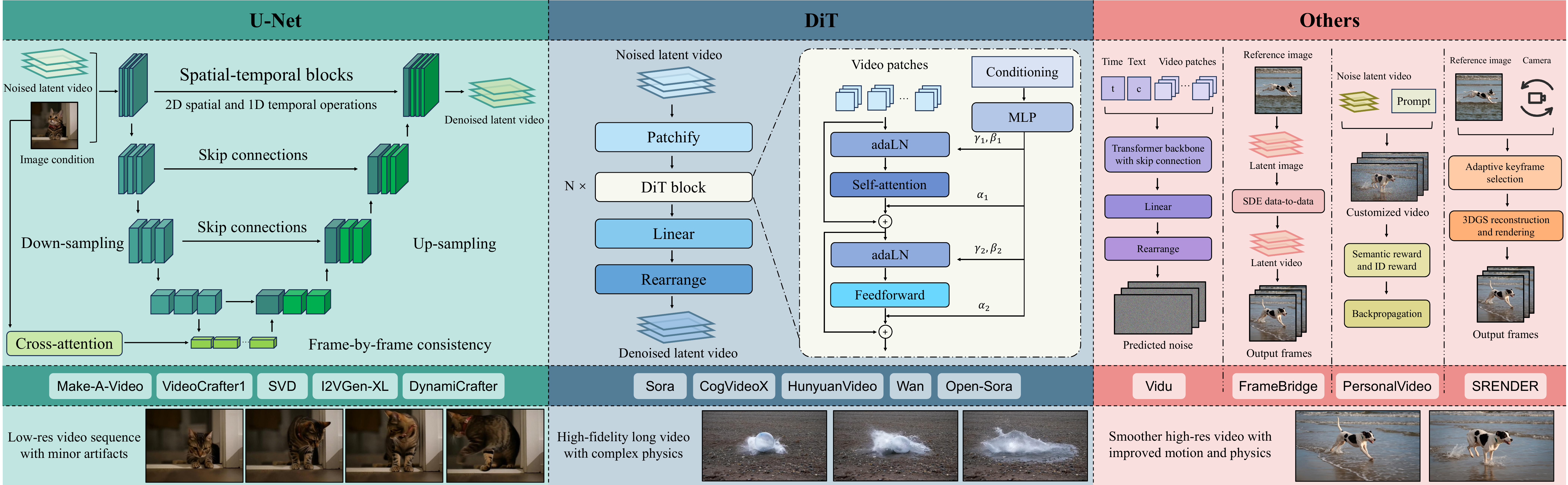}
    \caption{Illustration of   model architectures of diffusion I2V.}
    \label{fig:i2v-model}
\end{figure*}

For the basic I2V architecture, a spatiotemporal backbone combines spatial modeling, temporal interaction, and cross-attention to denoise video latents, which are then decoded into coherent video frames. Meanwhile, the reference image, diffusion timestep, and optional conditions are encoded into latent representations to guide video generation,   as illustrated in~\cref{fig:I2V-main}. 
The two most common I2V architectures are U-Net~\cite{ronneberger2015unet} and DiT~\cite{peebles2023scalable}, which are described in detail below (refer to~\cref{fig:i2v-model} for intuitive understanding):

\subsubsection{U-Net}
Early U-Net\footnote{In this paper, ``U-Net''   refers to a 2D U-Net~\cite{ronneberger2015unet} augmented with 1D temporal layers, essentially a pseudo-3D design as suggested in early works~\cite{singer2023make}, thus we adopt this convention for simplicity.}~\cite{singer2023make,blattmann2023stable,zhang2023i2vgenxl} paradigms extend pretrained \textit{text-to-image} (T2I) backbones (\eg, Stable Diffusion~\cite{Rombach2022stablediffusion}) with temporal modules to \textit{text-to-video} (T2V) models (also supports I2V). 
Specifically, the original 2D U-Net backbone is composed of spatial blocks distributed across the encoder, bottleneck, and decoder, while temporal blocks are additionally inserted to model inter-frame dynamics.  

A typical spatial block first applies 2D convolutions or residual blocks to each frame independently, producing frame-wise features $\mathbf{F}^{(l)} \in \mathbb{R}^{K \times h_l \times w_l \times d_l}$, where $l$ denotes the resolution level. 
Temporal modeling is then introduced by reshaping $\mathbf{F}^{(l)}$ along the frame axis and applying 1D temporal convolutions, temporal attention, or temporal Transformer layers across $K$ frames. 
As a result, each block factorizes video modeling into spatial processing within each frame and temporal interaction across frames. 
More concretely, a U-Net I2V block often follows the pattern: 
\begin{equation}
\mathbf{F}^{(l)}_{\text{out}}
=
\mathcal{T}^{(l)}
\Big(
\mathcal{S}^{(l)}(\mathbf{F}^{(l)}_{\text{in}})
\Big)     
\end{equation} 
where $\mathcal{S}^{(l)}$ denotes the spatial sub-block, usually implemented by 2D convolutional residual layers or spatial attention, and $\mathcal{T}^{(l)}$ denotes the temporal sub-block, which models inter-frame dependencies. 
Overall, the U-Net design is essentially a spatial image generator with temporal modules inserted between convolutional stages, and its core advantage lies in reusing mature T2I backbones with minimal structural modification.

\subsubsection{DiT}
More recently, \textit{Diffusion Transformer} (DiT)~\cite{yang2025megactor,xing2025motioncanvas,zhao2026phyrpr} architectures increasingly dominate the I2V field, which replace the convolutional encoder-decoder hierarchy with a Transformer~\cite{vaswani2017attention} that operates directly on video tokens. 
Instead of processing $\mathbf{Z}_t$ through downsampling and upsampling paths, DiT partitions the latent video into patches and projects them into token embeddings $\mathbf{X}_t$. 
Each token corresponds to a local region in space or spacetime, and positional encodings are added to preserve temporal order and spatial layout. 
The denoiser is then built as a stack of Transformer blocks, each consisting of layer normalization, multi-head self-attention, and a feed-forward network with residual connections, formulated as:
\begin{equation}
\tilde{\mathbf{X}}_t = \mathbf{X}_t + \mathrm{MSA}(\mathrm{LN}(\mathbf{X}_t))
\end{equation}
\begin{equation}
\mathbf{X}_{t}^{\mathrm{out}} = \tilde{\mathbf{X}}_t + \mathrm{MLP}(\mathrm{LN}(\tilde{\mathbf{X}}_t))
\end{equation}
where $\mathrm{LN}(\cdot)$ denotes layer normalization, $\mathrm{MSA}(\cdot)$ represents multi-head self-attention,  $\mathrm{MLP}(\cdot)$ denotes the feed-forward network,  $\tilde{\mathbf{X}}_t$ is the intermediate token representation after the self-attention layer, and $\mathbf{X}_{t}^{\mathrm{out}}$ is the output token representation of the current DiT block. 


%% file: sections/sec2_4_datasets_metrics.tex
\subsection{I2V Datasets}
\label{sec:datasets}
The development of I2V diffusion models has been supported by a growing collection of video datasets. Therefore, we review their evolution from the following two perspectives:

\subsubsection{Training Datasets}
As a key factor shaping generation quality, controllability, and generalization, the training data of I2V diffusion models broadly spans four primary categories (refer to~\cref{tab:training-datasets} for comparison details):

\begin{itemize}[label=\ding{117}]
\item \textbf{Text-Video Datasets.} 
WebVid-10M~\cite{bain2021frozen} emerges as a pivotal resource for early I2V research, providing 10M video clips paired with associated text captions, and is widely used to construct I2V training tuples by sampling the first frame (or a keyframe) as the image condition and the remaining frames as the target video. 
Such supervision proves critical for I2V models such as SVD~\cite{blattmann2023stable}, ConsistI2V~\cite{ren2024consisti2v}, Motion-I2V~\cite{shi2024motioni2v}, and DynamiCrafter~\cite{xing2024dynamicrafter}. 
As models grow more capable, larger datasets help alleviate this scale limitation. 
HD-VILA-100M~\cite{xue2022advancing} introduces 100M text-video pairs with 32.5-word captions and 720p resolution, while Panda-70M~\cite{chen2024panda70m} provides 70M semantically coherent clips with automatic captions and high semantic coverage. 
By enabling automatic conversion from text-video pairs to text-image-video training samples at scale, these datasets become valuable for training data-hungry I2V models, including DiT-based I2V frameworks such as Open-Sora~\cite{zheng2024opensora}  and HunyuanVideo~\cite{kong2024hunyuanvideo}.

\item \textbf{Text-Image Datasets.}  
Text-image datasets consist of large-scale image-caption pairs (rather than videos), which provide rich semantic and visual priors for learning appearance, composition, and text-image alignment. 
Strong visual priors from image models significantly improve video generation quality, leading I2V diffusion models like I2VGen-XL~\cite{zhang2023i2vgenxl}, VideoCrafter1~\cite{chen2023videocrafter1}, and MagicAnimate~\cite{xu2024magicanimate} to leverage large-scale text-image corpora such as LAION-5B~\cite{schuhmann2022laion} and LAION-COCO~\cite{LAIONCOCO600M} for initial training stages before acquiring temporal dynamics.  
In addition, high-quality text-image datasets like JourneyDB~\cite{sun2023journeydb} further provide diverse aesthetic styles and prompt-image alignment, which improve visual fidelity and controllability in downstream I2V generation.

\begin{table}[t]
\centering
\caption{Comparison of major I2V training datasets.}
\label{tab:training-datasets}
\renewcommand{\arraystretch}{1.2}
\scalebox{0.8}{
\begin{tabular}{>{\arraybackslash}m{2.7cm}|c|c|>{\centering\arraybackslash}m{3.6cm}}
\toprule[1.5pt] \rowcolor{highlightgray}
\textbf{Datasets} & \textbf{Scale} & \textbf{Type} & \textbf{Relevant I2V schemes} \\
\midrule[1.2pt]
WebVid-10M~\cite{bain2021frozen} & 10M pairs & Text-video & ConsistI2V~\cite{ren2024consisti2v} \\
WebVid-2M~\cite{bain2021frozen} & 2.5M pairs & Text-video & PixelDance~\cite{zeng2024make}, EDG~\cite{tian2025extrapolating} \\
HD-VILA-100M~\cite{xue2022advancing} & 100M clips & Text-video &  MoVideo~\cite{liang2024movideo} \\
Panda-70M~\cite{chen2024panda70m} & 70M pairs & Text-video &   Tora~\cite{zhang2025tora}, STIV~\cite{lin2025stiv} \\
OpenVid-1M~\cite{nan2024openvid} & 1M pairs & Text-video & Loopy~\cite{jiang2025loopy}, Dynamic-I2V~\cite{liu2025dynamic} \\
LAION-COCO~\cite{LAIONCOCO600M} & 600M pairs & Text-image & VideoCrafter1~\cite{chen2023videocrafter1} \\ 
LAION-400M~\cite{schuhmann2021laion} & 400M pairs & Text-image & I2VGen-XL~\cite{zhang2023i2vgenxl}  \\
LAION-5B~\cite{schuhmann2022laion} & 5B pairs & Text-image & Open-Sora Plan~\cite{lin2024opensora} \\
JourneyDB~\cite{sun2023journeydb} & 4M pairs & Text-image & EasyAnimate~\cite{xu2024easyanimate} \\ 
Hallo3~\cite{cui2025hallo3} & 134 hrs & Portrait &  FantasyTalking~\cite{wang2025fantasytalking} \\
HDTF~\cite{zhang2021flow} & 15.8 hrs & Portrait & EMO~\cite{tian2024emo}, Hallo~\cite{xu2024hallo} \\
VFHQ~\cite{xie2022vfhq} & 16K clips & Portrait &  AniPortrait~\cite{wei2024aniportrait}, Sonic~\cite{ji2025sonic} \\
CelebV-Text~\cite{yu2023celebv} & 70K clips & Portrait & Sonic~\cite{ji2025sonic}, MVPortrait~\cite{lin2025mvportrait} \\ 
CelebV-HQ~\cite{zhu2022celebvhq} & 35K clips & Portrait & VideoMaker~\cite{wu2024videomaker} \\ 
NeRSemble~\cite{kirschstein2023nersemble} & 7.5 hrs & Portrait & SkyReels-A1~\cite{qiu2025skyreels}  \\
RealEstate10K~\cite{zhou2018stereo} & 10K videos & House tours & CamI2V~\cite{zheng2024cami2v} \\
DL3DV-10K~\cite{ling2024dl3dv} & 10K videos & 3D scale & SRENDER~\cite{chen2026efficient}  \\  
AVSpeech~\cite{ephrat2018looking} & 4.7K hrs & Audio-visual & OmniAvatar~\cite{gan2025omniavatar}, EMO2~\cite{tian2025emo2} \\
ViViD~\cite{fang2024vivid} & 9.7K pairs & Video try-on &  DPIDM~\cite{li2025pursuing} \\ 
Proprietary dataset & - & Proprietary & CogVideoX~\cite{yang2025cogvideox}, Wan~\cite{wan2025open} \\ 
\bottomrule[1.5pt]
\end{tabular}}
\end{table}

\item \textbf{Portrait Datasets.}
Portrait datasets contain human-centric videos (\eg, face or full-body motions), often with high-quality identity, expression, and pose dynamics, making them suitable for portrait animation tasks that require subject consistency and fine-grained motion control. 
Specialized portrait datasets including HDTF~\cite{zhang2021flow}, VFHQ~\cite{xie2022vfhq}, and CelebV-HQ~\cite{zhu2022celebvhq} are employed in audio-driven animation methods like EMO~\cite{tian2024emo}, Hallo~\cite{xu2024hallo}, AniPortrait~\cite{wei2024aniportrait}, and MegActor-Sigma~\cite{yang2025megactor}. 

\item \textbf{Proprietary Datasets.} 
Leading commercial I2V systems increasingly rely on internal datasets whose characteristics remain largely undisclosed. HunyuanVideo~\cite{kong2024hunyuanvideo}, CogVideoX~\cite{yang2025cogvideox}, Wan~\cite{wan2025open}, and Seedance~\cite{gao2025seedance} all report training on proprietary data collections that likely exceed public datasets in both scale and curation quality. This trend creates a reproducibility gap in the field, as open-source efforts struggle to match the performance of models trained on inaccessible data.
\end{itemize}
 
In summary, I2V training datasets have evolved from general-purpose web-scale corpora to task-specialized and proprietary collections. 
This trend highlights that data curation and dataset composition are now as important as model architecture in advancing diffusion-based I2V generation. More details on I2V training datasets can be found in~\cref{tab:training-datasets}.

\subsubsection{Testing Datasets}
While training datasets determine what I2V diffusion models learn, testing datasets determine how their generation capabilities are assessed. 
For I2V evaluation, UCF-101~\cite{soomro2012ucf101} remains one of the most widely used test datasets, containing videos at 320$\times$240 resolution and 25 \textit{frames per second} (FPS). 
Although originally proposed for action recognition, it is commonly adopted in I2V studies to evaluate video realism and distribution quality, particularly via \textit{Fréchet Video Distance} (FVD)~\cite{unterthiner2018towards}. 
MSR-VTT~\cite{xu2016msr} is also frequently used in I2V evaluation, as it provides 10,000 videos with 20 human-written captions per video, enabling measurement of how well generated videos preserve semantic consistency. 
Beyond dataset-level benchmarks, VBench~\cite{huang2024vbench} has become a standardized evaluation suite for I2V and T2V generation, decomposing video quality into 16 dimensions spanning temporal quality (\eg, subject consistency, motion smoothness, dynamic degree), frame-wise quality (\eg, aesthetic quality, imaging quality), and video-condition consistency (\eg, object class, human action, spatial relationships). 
By using specialized pretrained evaluators validated against human preferences, VBench offers a more fine-grained and diagnostic assessment of I2V model performance.

\vspace{-0.5em}
\subsection{Evaluation Metrics}
Evaluating I2V generation remains challenging because it jointly assesses frame-level perceptual fidelity, temporal coherence, and semantic alignment with conditioning inputs.  
Specifically, 
\textit{Fréchet Video Distance} (FVD)~\cite{unterthiner2018towards} extends the \textit{Fréchet Inception Distance} (FID)~\cite{heusel2017gans} from images to videos by measuring the Fréchet distance between feature distributions of real and generated samples as follows:
\begin{equation}
\mathrm{FVD} = \|\mu_r - \mu_g\|^2 + \mathrm{Tr}(\Sigma_r + \Sigma_g - 2(\Sigma_r \Sigma_g)^{1/2})
\end{equation}
where $(\mu_r,\Sigma_r)$ and $(\mu_g,\Sigma_g)$ denote the mean and covariance of deep video features extracted from real and generated videos, respectively. In practice, these features are computed using a pretrained \textit{Inflated 3D ConvNet} (I3D)~\cite{carreira2017quo}, whose 3D spatiotemporal convolutions enable FVD to capture not only per-frame appearance but also motion and temporal dynamics. 
This distinguishes FVD from frame-based FID~\cite{heusel2017gans}, which evaluates images independently and may miss temporal artifacts that are  critical for I2V generation. 

\begin{table*}[t]
\centering
\caption{Comparison of evaluation metrics of I2V diffusion models.}
\label{tab:evaluation_metrics}
\renewcommand{\arraystretch}{1.2}
\scalebox{0.83}{\begin{tabular}{c|c|c|c|c}
\toprule[1.5pt] \rowcolor{highlightgray}
\textbf{Metrics} & \textbf{Evaluation paradigm} & \textbf{Evaluation reference data} & \textbf{Video \textit{vs.} frame-level evaluation}  & \textbf{Relevant I2V methods} \\
\midrule[1.2pt]
FVD~\cite{unterthiner2018towards} & Distribution-based & Ground-truth videos & Video-level & SVD~\cite{blattmann2023stable}, Lumiere~\cite{bartal2024lumiere} \\
IS~\cite{barratt2018note} & Classifier-based  &  N/A & Frame-level & PixelDance~\cite{zeng2024make}, W.A.L.T~\cite{gupta2024photorealistic}, Make-A-Video~\cite{singer2023make}\\
PSNR~\cite{fardo2016formal} & Pixel-level reconstruction  & Ground-truth video frames & Frame-level   & ViewCrafter~\cite{yu2024viewcrafter}, Pixel-to-4D~\cite{de2026pixel} \\
SSIM~\cite{nilsson2020understanding} &  Structural similarity  & Ground-truth video frames & Frame-level   & RCM~\cite{wang2026rcm}, Dreamvideo~\cite{wang2025dreamvideo} \\
LPIPS~\cite{zhang2018unreasonable} & Deep perceptual similarity  & Ground-truth video frames & Frame-level  & MagicAnimate~\cite{xu2024magicanimate}, Animate Anyone~\cite{hu2024animate} \\
VBench-I2V score~\cite{huang2024vbench} & Multi-dimension evaluation &  N/A  & Video-level and frame-level & FrameBridge~\cite{wang2025framebridge}, RealCam-I2V~\cite{li2025realcam}, Dynamic-I2V~\cite{liu2025dynamic} \\
\bottomrule[1.5pt] 
\end{tabular}}
\end{table*}

In addition, many works report perceptual image-quality in I2V models.
Specifically, \textit{Inception Score} (IS)~\cite{barratt2018note} is employed for quantifying generation quality of videos and diversity with a pretrained inception-based classifier~\cite{singer2023make,wang2025dreamvideo,gupta2024photorealistic}, formulated as:
\begin{equation}
\mathrm{IS}=\exp\left(\mathbb{E}_{v_k\sim \pi_g}\left[\mathrm{KL}\big(\pi(y|v_k)\,\|\,\pi(y)\big)\right]\right)
\end{equation}
where $v_k$ denotes the frame sampled from the distribution of generated videos $\pi_g$, $\pi(y|v_k)$ is the label distribution predicted by a pretrained inception classifier, $\pi(y)$ is the label distribution over generated samples, and $\mathrm{KL}(\cdot\|\cdot)$ denotes the KL divergence. 
Meanwhile, 
\textit{Peak Signal-to-Noise Ratio} (PSNR)~\cite{fardo2016formal}, \textit{Structural Similarity Index Measure} (SSIM), and \textit{Learned Perceptual Image Patch Similarity} (LPIPS)~\cite{zhang2018unreasonable} are used to evaluate the 
pixel-level fidelity of synthesized video frames against ground-truth frames~\cite{de2026pixel,yu2024viewcrafter,nilsson2020understanding,wang2026rcm,wang2025dreamvideo}. PSNR is formulated as follows: 
\begin{equation}
\mathrm{PSNR} = 10 \cdot \log_{10}\left( \frac{\mathrm{MAX}_v^2}{\mathrm{MSE}(v_{\text{syn}}, v_{\text{gt}})} \right)
\end{equation} 
where $v_{\text{syn}}$ and $v_{\text{gt}}$ denote the synthesized and ground-truth frames respectively, $\text{MAX}_v$ is the maximum pixel intensity, $\text{MSE}$ denotes the mean squared error. SSIM is calculated as: 
\begin{equation}
\mathrm{SSIM} = \frac{(2\mu_{\text{syn}}\mu_{\text{gt}} + C_1)(2\text{cov}(v_{\text{syn}}, v_{\text{gt}}) + C_2)}{(\mu_{\text{syn}}^2 + \mu_{\text{gt}}^2 + C_1)(\sigma_{\text{syn}}^2 + \sigma_{\text{gt}}^2 + C_2)}
\end{equation} 
where $\mu_{\text{syn}}, \mu_{\text{gt}}$, $\sigma_{\text{syn}}, \sigma_{\text{gt}}$, and $\text{cov}$ represent the mean, variance and covariance of local image patches, and $C_1$ and $C_2$ are small constants for numerical stability. 
Besides, LPIPS is formulated as follows: 
\begin{equation}
\mathrm{LPIPS} = \frac{1}{H_f \cdot W_f} \sum_{h_f,w_f} \left\| \phi(v_{\text{syn}})_{h_f,w_f} - \phi(v_{\text{gt}})_{h_f,w_f} \right\|_2^2
\end{equation}
where $H_f$ and $W_f$ represent the height and width of the feature map, 
$\phi(\cdot)$ denotes a pre-trained deep network for feature extraction. 
Beyond the aforementioned common image evaluation metrics, a customized VBench-I2V score~\cite{huang2024vbench,huang2025vbench++} tailored for the I2V scenario has also been proposed, which is designed to comprehensively quantify the quality of video generation from static images by evaluating temporal consistency, motion smoothness, visual fidelity, and semantic alignment with the input image~\cite{wang2025framebridge,li2025realcam,liu2025dynamic}. The VBench-I2V score is formulated as:
\begin{equation}
\mathrm{Score}_{\text{VBench-I2V}}  = \frac{1}{M} \sum_{m=1}^M w_m \cdot s_m(I_{\text{in}}, V_{\text{gen}})
\end{equation}
where $I_{\text{in}}$ denotes the input static image, $V_{\text{gen}}$ represents the generated video sequence, $M$ is the total number of I2V-specific evaluation dimensions (\eg, temporal consistency, motion realism), $w_m$ is the weight assigned to the $m$-th dimension, and $s_m(\cdot, \cdot)$ is the normalized score (ranged in $[0,1]$) of the $m$-th dimension for the pair of input image and generated video. 
In conclusion, an intuitive comparison for these I2V evaluation metrics is demonstrated in~\cref{tab:evaluation_metrics}.

%% file: sections/sec3_1_condition_type.tex
We provide a comprehensive review of I2V schemes and systematically categorize them as illustrated in~\cref{fig:I2V-taxmony}. 
To gain a deeper understanding, we analyze diffusion-based I2V approaches from the following two taxonomies: model architecture and training paradigm.

\subsection{Taxonomy by Model Architecture}
\label{sec:model_arch}
Based on model architectures, I2V schemes fall into three categories: U-Net based, DiT based, and other schemes. Comparison of representative schemes is provided in~\cref{tab:arch_comparison}. 

\subsubsection{U-Net Schemes} 
In early diffusion I2V, Make-A-Video~\cite{singer2023make} extends T2I models with temporal layers by adding a 1D convolution after each 2D convolution. 
VideoCrafter1~\cite{chen2023videocrafter1} applies this design to Stable Diffusion~\cite{Rombach2022stablediffusion} and further adopts joint image-video training to reduce concept forgetting. 
\textit{Stable Video Diffusion} (SVD)~\cite{blattmann2023stable} shifts focus from architectural novelty to data curation and training strategy, while I2VGen-XL~\cite{zhang2023i2vgenxl} adopts a cascaded U-Net pipeline, where a base U-Net models global semantics and motion and a refinement U-Net improves visual fidelity.

Subsequent U-Net based designs further refine how motion and conditions are injected. TRIP~\cite{zhang2024trip} introduces a dual-path U-Net with an image-noise shortcut and a residual branch for temporal modeling, I2V-Adapter~\cite{guo2024i2vadapter} adds lightweight cross-frame attention to a frozen video U-Net for identity propagation, and AID~\cite{xing2024aid} extends pretrained I2V U-Nets with temporal and spatial adapters for instruction-guided prediction. 
Recent variants continue this trend by attaching stronger motion or structure priors to the U-Net backbone, such as dual-direction motion fine-tuning in VTG~\cite{yang2025vtg}, explicit 3D guided generation in I2V3D~\cite{zhang2025i2v3d}, and motion-aware feature injection from a homologous diffusion network in Moaw~\cite{zhang2026moaw}. 
Overall, the U-Net line evolves from simple temporal extension to increasingly modular designs that separate appearance preservation, motion modeling, and conditional control.


\subsubsection{DiT Schemes} 
In I2V diffusion models, the \textit{Diffusion Transformer} (DiT)~\cite{peebles2023scalable,yan2026animeagent,liu2026tele,wu2026consid,li2026rerope,team2026fsvideo,shao2024human4dit,song2025makeanything,zhang2025packing}   replaces the U-Net denoiser with a Transformer~\cite{vaswani2017attention}, thereby modeling video generation in a unified token space.  
Specifically, early DiT-based schemes establish the backbone from the perspectives of attention design, multimodal fusion, and scaling. 
Open-Sora~\cite{zheng2024opensora} proposes \textit{Spatial-Temporal Diffusion Transformer} (ST-DiT), which decouples spatial and temporal attention, while EasyAnimate~\cite{xu2024easyanimate} follows this line and improves long-video efficiency with hybrid window attention. CogVideoX~\cite{yang2025cogvideox} introduces an expert transformer with expert adaptive LayerNorm for stronger text-video fusion. 
HunyuanVideo~\cite{kong2024hunyuanvideo} further shows that large DiT-based video diffusion models scale effectively with strong training efficiency, and Wan~\cite{wan2025open} validates the same Transformer paradigm under flow matching with both efficiency-oriented and high-capacity variants. 

Building on these general backbones, recent I2V oriented DiT models mainly evolve toward richer condition injection and stronger task specialization. 
Dynamic-I2V~\cite{liu2025dynamic} and OmniV2V~\cite{liang2025omniv2v} extend DiT into unified conditional generators by injecting multi-modal instructions or dynamic signals into the Transformer. 
In human-centric generation, HumanDiT~\cite{gan2025humandit} uses pose-guided DiT blocks with prefix latent reference conditioning, UniAnimate~\cite{wang2025unianimate} unifies the reference image, pose guidance, and noisy video in a shared video diffusion backbone, and DreamVVT~\cite{zuo2025dreamvvt} adopts a stage-wise DiT framework in which keyframe try-on synthesis provides appearance guidance for subsequent video generation. 
This specialization trend is further exemplified by HunyuanVideo-Avatar~\cite{chen2025hunyuanvideo}, which builds on the large-scale HunyuanVideo line and equips the multimodal DiT backbone with character image injection, audio emotion modeling, and face-aware audio adapters for multi-character animation. 
Related extensions continue to treat DiT as a flexible multi-modal host, including FlexiMMT~\cite{li2026let} with mask-constrained attention for multi-object multi-motion transfer and SkyReels-V4~\cite{chen2026skyreels} with a dual-stream MM-DiT for joint video-audio generation, inpainting, and editing. In conclusion, DiT-based I2V approaches evolve from general scalable video Transformers into condition-centric and task-aware generators, which makes them more flexible for multi-modal control and more suitable for large-scale   modeling.

\subsubsection{Other Schemes}
Beyond U-Net and DiT based I2V schemes, several recent works explore alternative design paradigms. 
Vidu~\cite{bao2024vidu} adopts a U-ViT backbone to improve long-video consistency and scalability, representing a Transformer based video generator that does not follow the standard DiT. 
FrameBridge~\cite{wang2025framebridge} departs more fundamentally from diffusion I2V by replacing noise-to-data generation with a data-to-data bridge process, which better matches the frame-to-frame nature of image animation. PersonalVideo~\cite{li2024personalvideo} focuses on identity-preserving video customization through an isolated identity adapter and direct video-level supervision, emphasizing non-intrusive personalization rather than backbone redesign.  SRENDER~\cite{chen2026efficient} instead combines sparse diffusion with 3D reconstruction and rendering, generating only a small set of key frames and synthesizing the full video through geometric rendering. 
Accordingly, these methods provide complementary perspectives  by exploring alternative generation paradigms in I2V. 
Based on the above discussion, we derive the following insight: 

\begin{remark}[Remark I (\textit{Architecture Comparison})]
U-Net often saturates as scale increases, whereas DiT continues to benefit from larger models while supporting arbitrary spatiotemporal layouts through token-based representations. This indicates that DiT is better aligned with the long-term direction of I2V, where scalability, condition flexibility, and multimodal unification matter as much as generation quality itself.
\end{remark}

%% file: sections/sec3_2_training_paradigm.tex
\begin{table*}[!t]
\centering
\renewcommand{\tabularxcolumn}[1]{m{#1}} 
\setlength{\tabcolsep}{3.5pt} 
\setlength{\aboverulesep}{0pt} 
\setlength{\belowrulesep}{0pt}
\renewcommand{\arraystretch}{1.2}

\definecolor{headBlue}{HTML}{DAE8FC}  
\definecolor{headYellow}{HTML}{FFF2CC} 
\definecolor{rowGray}{HTML}{F5F5F5}    
\definecolor{grpGray}{HTML}{EBEBEB}    

\caption{Comparison of representative I2V architectures (``N/M" denotes ``Not mentioned"). } 
\label{tab:arch_comparison}

\begin{tabularx}{\linewidth}{>{\centering\arraybackslash}m{2.3cm} >{\centering\arraybackslash}m{0.5cm} >{\centering\arraybackslash}m{1.25cm} >{\centering\arraybackslash}m{1.9cm} >{\centering\arraybackslash}m{1cm} X}
\toprule 

\cellcolor{headBlue}\textbf{I2V model} & 
\cellcolor{headBlue}\textbf{Year} & 
\cellcolor{headBlue}\textbf{Venue} & 
\cellcolor{headBlue}\textbf{Arch.} & 
\cellcolor{headBlue}\textbf{Param.} & 
\cellcolor{headYellow}\textbf{Key features} \\
\midrule

\multicolumn{6}{c}{\cellcolor{grpGray}\textbf{U-Net~\cite{ronneberger2015unet}-based schemes}} \\
\hline 
Make-A-Video~\cite{singer2023make} & 2023 & ICLR & U-Net & N/M & It enables I2V through spatiotemporally factorized diffusion models.\\
\rowcolor{rowGray} VideoCrafter1~\cite{chen2023videocrafter1} & 2023 & arXiv & U-Net & N/M & It animates reference images while preserving their content and structures. \\
SVD~\cite{blattmann2023stable} & 2023 & arXiv & U-Net & 1.52B & It establishes motion priors through  data curation and three-stage training. \\
\rowcolor{rowGray} I2VGen-XL~\cite{zhang2023i2vgenxl} & 2023 & arXiv & U-Net & N/M & It uses a cascaded pipeline to decouple semantic coherence and resolution refinement.\\
LFDM~\cite{ni2023conditional} & 2023 & CVPR & U-Net & N/M & It synthesizes optical flow sequence in the latent space to warp the given image.\\
\rowcolor{rowGray} 
Lumiere~\cite{bartal2024lumiere} & 2024 & SA & ST U-Net & N/M & It generates videos in one pass with spatiotemporal downsampling and upsampling.\\
DynamiCrafter~\cite{xing2024dynamicrafter} & 2024 & ECCV & U-Net & N/M & It uses dual-stream injection combining query transformer and  noise concatenation.\\
\rowcolor{rowGray} 
ConsistI2V~\cite{ren2024consisti2v} & 2024 & TMLR & U-Net & 1.7B & It enhances consistency via spatiotemporal attention and low-frequency initialization.\\
Motion-I2V~\cite{shi2024motioni2v} & 2024 & arXiv & U-Net & N/M & It adopts motion field predictor and motion-augmented temporal attention.\\
\midrule 
 
\multicolumn{6}{c}{\cellcolor{grpGray}\textbf{DiT~\cite{peebles2023scalable}-based schemes}} \\
\hline
W.A.L.T~\cite{gupta2024photorealistic} & 2024 & ECCV & DiT & 3B & It utilizes encoder for compression and window attention for modeling.\\
\rowcolor{rowGray} 
Open-Sora~\cite{zheng2024opensora} & 2024 & arXiv & ST-DiT & 1.1B & It decouples spatial and temporal attention via a compressive 3D autoencoder. \\
HunyuanVideo~\cite{kong2024hunyuanvideo} & 2024 & arXiv & DiT & 13B & It introduces an open-source model utilizing progressive scaling. \\
\rowcolor{rowGray} 
CogVideoX~\cite{yang2025cogvideox} & 2025 & ICLR & DiT & 2B/5B & It combines a 3D causal VAE and expert transformer via AdaLN for modality fusion. \\
Wan~\cite{wan2025open} & 2025 & arXiv & DiT & 1.3B/14B & It utilizes a spatio-temporal VAE and scalable pre-training.\\
\rowcolor{rowGray} 
EasyAnimate~\cite{xu2024easyanimate} & 2025 & ACM MM & DiT & N/M & It integrates slice VAE and hybrid motion module with temporal and global attention. \\
Dynamic-I2V~\cite{liu2025dynamic} & 2025 & arXiv & DiT & 7B & It incorporates MLLMs via a conditional adapter to encode vision and text. \\
\rowcolor{rowGray} 
OmniV2V~\cite{liang2025omniv2v} & 2025 & arXiv & DiT & 13B & It injects dynamic content injection with LLaVA-based instructions.\\
HumanDiT~\cite{gan2025humandit} & 2025 & arXiv & DiT & 5B & It adopts a prefix-latent strategy and pose adapter.\\
\rowcolor{rowGray} 
UniAnimate~\cite{wang2025unianimate} & 2025 & SCIS  & DiT & 14B & It fuses a 3D convolution pose encoder via LoRA and appearance concatenation. \\
\mbox{ {\fontsize{6pt}{7pt}\selectfont HunyuanVideo-Avatar~\cite{chen2025hunyuanvideo}} } & 2025 & arXiv & MM-DiT & 13B & It embeds an audio emotion module and face-aware audio adapter.\\
\rowcolor{rowGray} StableAvatar~\cite{tu2025stableavatar} & 2025 & arXiv & DiT & 1.3B & It integrates audio adapter and guidance mechanism to synthesize avatar videos.\\
SANA-Video~\cite{chen2025sana} & 2025 & arXiv & DiT & 2B & It utilizes block linear attention and constant-memory state for minute-long videos.\\
\rowcolor{rowGray}

\mbox{ {\fontsize{6pt}{7pt}\selectfont MegActor-Sigma~\cite{yang2025megactor}}} & 2025 & AAAI & DiT & 1.4B & It integrates audio-visual controls for portrait animation. \\
FlexiMMT~\cite{li2026let} & 2026 & arXiv & MM-DiT & N/M & It decouples cross-object motion via object-specific attention and mask propagation.\\
\rowcolor{rowGray}
SkyReels-V4~\cite{chen2026skyreels} & 2026 & arXiv & MM-DiT & N/M & It integrates MLLMs to inject fine-grained visual guidance via in-context learning.\\
\midrule  

\multicolumn{6}{c}{\cellcolor{grpGray}\textbf{Other schemes}} \\
\hline
Vidu~\cite{bao2024vidu} & 2024 & arXiv & U-ViT & N/M & It integrates U-ViT for scalable, single-pass high-definition video generation.\\
\rowcolor{rowGray} FrameBridge~\cite{wang2025framebridge} & 2025 & ICML & Data-to-data  & - & It proposes a data-to-data bridge via SNR-aligned tuning and neural priors. \\
PersonalVideo~\cite{li2024personalvideo} & 2025 & ICCV & Isolated adapter & - &It integrates ID and semantic rewards via supervision and  prompt augmentation.\\
\rowcolor{rowGray} SRENDER~\cite{chen2026efficient} & 2026 & arXiv & Reconstruction
 & - & It accelerates synthesis by lifting adaptive keyframes to 3D for intermediate rendering. \\
\bottomrule 
\end{tabularx}
\end{table*}

\subsection{Taxonomy by Training Paradigm}
\label{sec:training_paradigm}

Training paradigms for I2V balance model capability, training efficiency, and practical deployment. We categorize existing approaches into three paradigms: training-free methods, multi-stage training, and fine-tuning adapters, as follows:


\subsubsection{Training-free}
Training-free methods leverage pretrained models through inference-time manipulation without parameter updates. 
Early approaches such as SD~\cite{liu2023pilife}, I4VGen~\cite{guo2024i4vgen}, and TI2V-Zero~\cite{ni2024ti2v} have explored prompt engineering and feature injection, while SG-I2V~\cite{namekata2025sgi2v} adds trajectory control and PhysGen~\cite{liu2024physgen} integrates physics simulation. 
Recent efficiency-oriented methods exploit inherent sparsity in video diffusion transformers. 
Specifically, SVG~\cite{xi2025sparse} and SVG2~\cite{yang2025sparse} improve efficiency by exploiting spatial-temporal sparsity in video attention, with the former separating attention heads and the latter   enhancing sparse execution. UniCP~\cite{sun2025unicp} instead improves efficiency through a unified design that combines dynamic feature caching with attention pruning.

\subsubsection{Single-stage Training}
Under single-training paradigm, existing I2V works usually insert image-aware mechanisms into the same backbone.  
One line of work realizes this design by modifying the denoising process itself, such as masking in SEINE~\cite{chen2024seine}, spatiotemporal attention in ConsistI2V~\cite{ren2024consisti2v}, and residual prediction in TRIP~\cite{zhang2024trip}. 
Meanwhile, related variants involve  adapter insertion~\cite{guo2024i2vadapter}, full space-time generation~\cite{bartal2024lumiere}, bridge modeling~\cite{wang2025framebridge}, transition  diffusion~\cite{yang2025vtg}, or context packing~\cite{zhang2025framepack}.  
A second line keeps the same single model but injects structured conditions, including camera-aware control in CamCo~\cite{xu2024camco} and CamI2V~\cite{zheng2024cami2v}, geometry-aware modeling in Diffusion4D~\cite{liang2024diffusion4d}, and trajectory guidance in Levitor~\cite{wang2025levitor}.  
Further schemes introduce motion representations, positional control, and 3D-aware constraints~\cite{xing2025motioncanvas,gu2025diffusion,li2025realcam,watson2024controlling,chen2026efficient,wu2024draganything,li2026let,li2026rerope,sun2024dimensionx}. 
The same formulation is also extended to fidelity-oriented generation in AtomoVideo~\cite{gong2024atomovideo} and MagDiff~\cite{zhao2024magdiff}, and to customization or adaptation in VideoMaker~\cite{wu2024videomaker}, PersonalVideo~\cite{li2024personalvideo}, and AID~\cite{xing2024aid}, with similar extensions in creation settings~\cite{zeng2024make,liu2025dynamic,jiang2025vace,wu2025customcrafter}. 
Human-centered variants  incorporate character decomposition in MIMO~\cite{men2024mimo}, pose guidance in MimicMotion~\cite{zhang2025mimicmotion}, and identity preservation in StableAnimator~\cite{tu2025stableanimator}. 
Based on this, some schemes also combine human pose~\cite{gan2025humandit,li2025pursuing,hu2025animate,wang2025unianimate,tan2025animatex,wang2025unianimatedit}, audio conditions~\cite{tu2025stableavatar}, and identity constraints~\cite{wu2026consid}.

\subsubsection{Multi-stage Training}
Multi-stage training has become a central optimization paradigm in I2V diffusion models, progressively decomposing learning into sequential phases to stabilize training and enhance controllability. 
Specifically, 
SVD~\cite{blattmann2023stable} follows a two-stage adaptation strategy for I2V generation. It first adapts a pretrained T2V model to image-conditioned generation by replacing text input with CLIP image features and a noise-augmented frame latent, and then fine-tunes it on high-resolution video data to improve visual quality and temporal consistency. 
This design proves effective for audio-driven animation tasks. 
For example, Hallo~\cite{xu2024hallo}, Hallo2~\cite{cui2024hallo2}, AniPortrait~\cite{wei2024aniportrait}, Loopy~\cite{jiang2025loopy}, and EchoMimic~\cite{chen2025echomimic} model static facial appearance before introducing audio conditioning, while pose-driven approaches including UniAnimate~\cite{wang2025unianimate}, Champ~\cite{zhu2024champ}, and MimicMotion~\cite{zhang2025mimicmotion} learn body appearance prior to pose-to-motion mapping.  
Relevant designs also appear in Animate Anyone~\cite{hu2024animate}, Motion-I2V~\cite{shi2024motioni2v}, and Tora~\cite{zhang2025tora}.  
Thereafter, to address quality bottlenecks, three-stage training strategies~\cite{tian2024emo,xing2024dynamicrafter,lin2025omnihuman,zhang2024pia,zhang2024moonshot,wang2024boximator,wang2024vexpress,wang2024loopanimate,tian2025extrapolating,luo2025dreamactor} are proposed to achieve high-resolution enhancement, motion calibration, or modality-specific conditioning.   
At larger scales, foundation models further extend this paradigm into curriculum-style pipelines that scale resolution, aspect ratio, duration, and conditioning complexity to mitigate gradient instability, catastrophic forgetting, and dataset-scale constraints, as exemplified by Open-Sora~\cite{zheng2024opensora}, HunyuanVideo~\cite{kong2024hunyuanvideo}, SkyReels-V2~\cite{chen2025skyreels}, Make-A-Video~\cite{singer2023make}, EasyAnimate~\cite{xu2024easyanimate}, MarDini~\cite{liu2024mardini}, Reducio-DiT~\cite{tian2025reducio}, Waver~\cite{zhang2025waver}, and Seedance~\cite{gao2025seedance}. 
Overall, staged optimization emerges as a practical and scalable solution for balancing visual fidelity, temporal consistency, and controllability in I2V systems.  
From the above analysis, we obtain the following insight: 



\begin{remark}[Remark II (\textit{Optimization Shapes Representation})]
In I2V, the optimization scheme does not only affect how a model is trained; it determines whether motion is treated as an inference-time prior, a jointly learned dependency, or a progressively isolated objective. As a result, appearance preservation, temporal evolution, and controllability are not merely balanced differently, but are structurally encoded in different ways.
\end{remark}

%% file: sections/sec4_1_2_encoding_temporal.tex
\begin{figure*}
    \centering
    \includegraphics[width=0.93\linewidth]{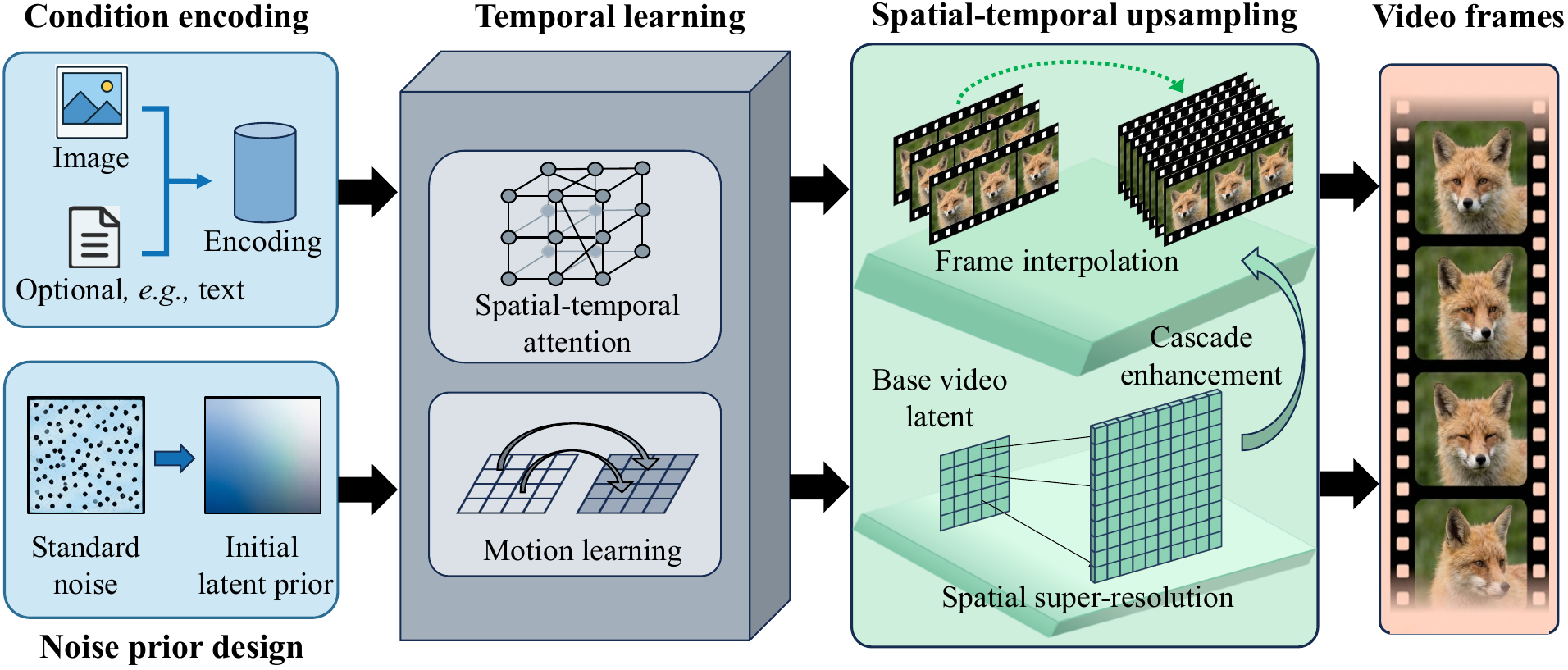}
    \caption{Illustration of four key technologies in I2V diffusion process.}
    \label{fig:key-design}
\end{figure*}

This section reviews four core designs in I2V models as demonstrated in~\cref{fig:key-design}, which govern how input conditions are represented and fused, how motion and temporal consistency are modeled, how diffusion trajectories are initialized and guided, and how video quality is  enhanced. 
\vspace{-0.5em}
\subsection{Condition Encoding}
\label{sec:condition_encoding}

Injecting conditioning information into I2V diffusion models requires balancing fidelity to the input image with controllability through conditions~\cite{zhang2024trip,ni2024ti2v,xing2024dynamicrafter,ren2024consisti2v,chen2023videocrafter1,shi2024motioni2v}.  
This section examines how I2V models encode conditions such as image and text into video generation process.  

\subsubsection{Image Condition Encoding}
The standard image-conditioning strategy concatenates the encoded image latent with the noisy video latent along the channel dimension~\cite{blattmann2023stable,girdhar2024emu,zeng2024make}, which can be formulated as follows:
\begin{equation}
z_{\text{input}} = \mathrm{concat}(z_{\text{img}}, z_t) \in \mathbb{R}^{(T/r_t) \times (H/r_s) \times (W/r_s) \times (C+C_{\text{img}})}
\end{equation} 
where $z_{\text{img}}$ is the encoded image latent, $z_t$ denotes the noisy video latent, $r_t$ and $r_s$ denote the temporal and spatial compression ratios, respectively, and $C$ and $C_{\text{img}}$ denote the channel dimensions of $z_t$ and $z_{\text{img}}$. The channel-wise concatenation of $z_{\text{img}}$ and $z_t$ forms $z_{\text{input}}$, which is then fed into the denoising network. 
This provides pixel-level guidance, though it requires modifying the first convolutional layer, typically initialized to zero to preserve pretrained weights.

Additionally, certain schemes~\cite{hu2024animate,chen2025hunyuanvideo,xing2024dynamicrafter} incorporate image conditions by injecting reference information  at different representation levels. 
Concretely, Animate Anyone~\cite{hu2024animate} introduces ReferenceNet to extract multi-scale spatial features from the image. HunyuanVideo-Avatar~\cite{chen2025hunyuanvideo} replaces additive conditioning with direct latent replacement for the first frame, eliminating the need for learned projection layers. DynamiCrafter~\cite{xing2024dynamicrafter} employs dual-stream injection where one path uses a query transformer to project CLIP image features into a text-aligned representation space.

\subsubsection{Text Condition Encoding}
Text conditioning provides semantic control over the dynamics and content of generated videos. 
A common formulation encodes the prompt $\tau$ into text tokens, \ie, $e_\tau = \mathcal{T}(\tau)$, where $\mathcal{T}(\cdot)$ denotes a text encoder such as T5~\cite{raffel2020exploring}.    
The token embeddings $e_\tau$ are injected into the denoising network through cross-attention layers:
\begin{equation}
\mathrm{Attn}(z,e_\tau)=\mathrm{softmax}\left(
\frac{(z W_Q)(e_\tau W_K)^\top}{\sqrt{d}}
\right) (e_\tau W_V) 
\end{equation}
where $z$ denotes the video latent feature, $W_Q, W_K, W_V$ are learnable projection matrices, and $d$ is the feature dimension. 
The core of text encoding in I2V is not only to obtain prompt embeddings, but to align text with image and video representations so that the prompt can guide both appearance preservation and motion synthesis. 
DynamiCrafter~\cite{xing2024dynamicrafter} makes this explicit by projecting the input image into a text-aligned feature space with a query transformer, thereby bridging image guidance and text-conditioned video priors. 
Emu Video~\cite{girdhar2024emu} instead factorizes generation into T2I and I2V stages, which preserves strong text-driven appearance generation before temporal extension. 
At larger scales, Open-Sora and Open-Sora Plan~\cite{zheng2024opensora,lin2024opensora} show that DiT based backbones can absorb text conditioning smoothly at billion-parameter scale, while industrial systems mainly refine the fusion mechanism itself, such as expert adaptive LayerNorm in CogVideoX~\cite{yang2025cogvideox}, dual-stream text-video processing in HunyuanVideo~\cite{kong2024hunyuanvideo}, and continued scaling in Wan~\cite{wan2025open} and Seedance~\cite{gao2025seedance}.

\subsubsection{Multi-modal Encoding}

The core of multi-modal encoding in I2V is to project heterogeneous conditions into a shared conditional space and then inject them into the denoising backbone~\cite{lin2025omnihuman,zhang2024moonshot,niu2024mofa}, formulated as: 
\begin{equation}
\mathbf{u}_m=\mathcal{E}_m(\mathbf{s}_m), \quad \tilde{\mathbf{u}}_m=\gamma_m \cdot \mathcal{P}_m(\mathbf{u}_m) 
\end{equation} 
\begin{equation}
\mathbf{g}=\mathcal{F}\big(\{\tilde{\mathbf{u}}_m\}_{m=1}^{M}\big)
\end{equation}
where $\mathbf{s}_m$ denotes the $m$-th input condition, $\mathcal{E}_m(\cdot)$ is its modality-specific encoder, $\mathcal{P}_m(\cdot)$ projects it into a common control space, $\gamma_m$ is an adaptive strength factor, and $\mathcal{F}(\cdot)$ aggregates all projected conditions into a unified guidance representation $\mathbf{g}$, which is then injected into the denoiser $\boldsymbol{\epsilon}_{\theta}(\mathbf{Z}_t,t;\mathbf{g})$.  
Under this view, the key challenge is not simply adding more modalities, but aligning their semantics, granularity, and influence so that different conditions remain compatible rather than interfering with one another.

Existing methods mainly differ in how they instantiate $\mathcal{P}_m(\cdot)$, $\gamma_m$, and $\mathcal{F}(\cdot)$. OmniHuman-1~\cite{lin2025omnihuman} emphasizes mixed-modality training with adaptive condition strength, making modality balancing the central design choice. MoonShot~\cite{zhang2024moonshot} builds a multimodal video block with decoupled cross-attention, so image and text conditions are fused in a unified yet explicitly separated manner. MOFA-Video~\cite{niu2024mofa} adopts a motion-centric design by converting different controls into motion field adaptations, which effectively uses motion as the shared intermediate interface. MegActor-Sigma~\cite{yang2025megactor} further enables flexible mixed-modal joint control in a diffusion transformer, making cross-modal interaction more explicit within the backbone. PhysGen~\cite{liu2024physgen} and PhyRPR~\cite{zhao2026phyrpr} extend the notion of modality beyond semantics by treating physical simulation or physical reasoning as structured guidance, while VACE~\cite{jiang2025vace} abstracts these inputs into a universal conditioning architecture that supports multiple generation and editing tasks within one framework. Taken together, these methods suggest that multi-modal encoding in I2V is evolving from simple condition aggregation to structured condition orchestration, and their main difference lies in what they choose as the common interface for alignment.

\vspace{-0.8em}
\subsection{Temporal Dynamics Learning}
\label{sec:motion_modeling}
Learning temporal dynamics presents challenges from spatial image generation. 
Motion patterns exhibit complex dependencies across time, requiring I2V models to capture both short frame-to-frame transitions and long temporal coherence.

\subsubsection{Temporal Attention} 
A straightforward approach to temporal modeling is to 
extend spatial self-attention along the temporal dimension~\cite{chen2023videocrafter1,shi2024motioni2v}. At each spatial position $n$, temporal attention aggregates information across all frames, learning how objects evolve across frames. 
Given spatiotemporal features $z \in \mathbb{R}^{K \times N \times d}$, where $K$ denotes the number of frames and $N$ is the number of spatial tokens in each frame, temporal attention at spatial position $n$ operates on the corresponding temporal feature sequence $z_n \in \mathbb{R}^{K \times d}$. 
The temporal attention is formulated as:
\begin{equation} 
\mathrm{Attn}_t(z_n)= \mathrm{softmax}\left(
\frac{(z_n W'_Q)(z_n W'_K)^\top}{\sqrt{d}}
\right)
(z_n W'_V) 
\end{equation}
where $W'_Q, W'_K, W'_V$ are matrices for temporal attention.

Specifically, PixelDance~\cite{zeng2024make} largely follows this design, and further strengthens inter-frame dependency modeling by introducing bidirectional self-attention in the temporal attention layers.  
Furthermore, EDG scheme~\cite{tian2025extrapolating}  treats temporal attention as a condition-aware motion controller rather than a pure frame-mixing operator.  
A closely related yet more conservative design is to inherit temporal attention from a pretrained backbone without substantially modifying its formulation. 
DreamVVT~\cite{zuo2025dreamvvt}, W.A.L.T.~\cite{gupta2024photorealistic}, and 4DiM~\cite{watson2024controlling} absorb temporal modeling into unified Transformer self-attention over spatiotemporal tokens or a joint space-time architecture,  so temporal interaction becomes part of a more global attention mechanism rather than an explicitly separated temporal branch.
Simultaneously, to improve efficiency, Open-Sora~\cite{zheng2024opensora} introduces ST-DiT with spatial-temporal attention that decouples spatial and temporal dimensions. 
As for long video synthesis, causal attention masks prevent the model from accessing future information. Therefore, CausVid~\cite{yin2025causvid} employs such masks to restrict attention to past frames only, enabling efficient streaming generation.

\subsubsection{Motion Learning}
While attention mechanisms learn motion patterns implicitly through data, explicit motion representations provide stronger inductive biases. 
LFDM~\cite{ni2023conditional} learns motion through latent flow diffusion, predicting optical flow fields that warp previous frames, formulated as:
\begin{equation}
\hat{z}_{k+1} = \mathcal{W}(z_k, g_{k \to k+1})
\end{equation}
where $z_k$ denotes the latent representation of the $k$-th frame, $\hat{z}_{k+1}$ is the warped latent representation aligned to frame $k+1$, $\mathcal{W}$ denotes a warping operation, and $g_{k \to k+1}$ represents the optical flow from frame $k$ to $k+1$. 
This decomposition separates content from motion, enabling independent control over content and motion dynamics. 
LFDM~\cite{ni2023conditional} further decomposes video generation into distinct content and motion diffusion processes to ensure alignment. 
Besides, to achieve trajectory control, Tora~\cite{zhang2025tora} introduces a 3D motion VAE that compresses sparse user-specified trajectories into a compact latent representation, enabling the model to generate videos following specified object paths.  
Implicit motion modeling offers an alternative by learning motion priors from video data. Specifically, SVD~\cite{blattmann2023stable} implements a motion bucket mechanism that conditions generation on motion scores derived from optical flow analysis. MoVideo~\cite{liang2024movideo} extends this approach with motion-aware video generation that captures diverse motion patterns across scenes, while LMP~\cite{chen2025lmp} demonstrates that these learned motion priors transfer effectively to zero-shot settings.


%% file: sections/sec4_3_5_noise_upsample_ctrl.tex
\vspace{-0.5em}
\subsection{Noise Prior Design}
\label{sec:noise_priors}
The initialization of noise in I2V diffusion models shapes the quality and consistency of generated animations. 
While traditional diffusion models sample noise independently for each frame, this independence creates temporal inconsistencies and weakens the relationship between the conditioning image and subsequent frames~\cite{ren2024consisti2v,zhang2024trip}. 

Early models like SVD~\cite{blattmann2023stable} implicitly incorporate shared noise through latent concatenation, creating coupling at the feature level.
Thereafter, TRIP~\cite{zhang2024trip} introduces explicit dual-path noise decomposition that separates temporal noise into conceptually distinct components. 
The shortcut path computes an image noise prior as follows:
\begin{equation}
    \epsilon_{t}^{i\to 1}=\frac{z_{t}^{i}-\sqrt{\bar{\alpha}_{t}}z_{0}^{1}}{\sqrt{1-\bar{\alpha}_{t}}}
\end{equation}
where $z_t^i$ denotes the noisy latent of the $i$-th frame at diffusion timestep $t$, 
$z_0^1$ is the clean latent of the first frame serving as a content reference, 
and $\bar{\alpha}_t = \prod_{s=1}^{t} \alpha_s$ is the cumulative noise schedule coefficient.
The formulation follows the standard diffusion reparameterization and extracts a frame-wise noise prior aligned with the first-frame content. 
Complementing this decomposition approach, ConsistI2V~\cite{ren2024consisti2v} proposes low-frequency noise initialization to favor smooth temporal transitions, recognizing that high-frequency noise components often introduce rapid, uncorrelated changes across frames. 
The above approaches share the assumption that generation proceeds from pure noise to clean data. 
However, FrameBridge~\cite{wang2025framebridge} challenges this by reformulating I2V generation as a data-to-data transformation, which uses the input image latent as the generation prior and establishes a bridge process that transforms this representation into video frames. 
Since the conditioning image already provides appearance information, regenerating all content from noise wastes computational resources and risks losing fine-grained details. 
 
\vspace{-0.5em}
\subsection{Spatial and Temporal Upsampling}
\label{sec:upsampling} 
In I2V generation, spatial and temporal upsampling improve visual fidelity by increasing spatial resolution and temporal density in a coarse-to-fine manner~\cite{zhang2023i2vgenxl,blattmann2023stable,bartal2024lumiere}.  
A unified spatial and temporal upsampling is to learn a refinement mapping from a coarse video representation to a denser one:
\begin{equation}
\mathcal{Y}^{\uparrow} = \mathcal{U}_{\psi}(\mathcal{Y}^{\downarrow}, \boldsymbol{\kappa}) 
\end{equation}
where $\mathcal{Y}^{\downarrow} \in \mathbb{R}^{T_c \times H_c \times W_c \times D}$ denotes a coarse video representation with reduced temporal length $T_c$ and spatial resolution $H_c \times W_c$, $\boldsymbol{\kappa}$ denotes auxiliary guidance such as the reference image, text prompt, or motion cues, and $\mathcal{U}_{\psi}(\cdot)$ is an upsampling module that produces the refined video representation $\mathcal{Y}^{\uparrow} \in \mathbb{R}^{T_r \times H_r \times W_r \times D}$, with $T_r\textgreater T_c$, $H_r\textgreater H_c$, and $W_r\textgreater  W_c$. 
In particular, SVD~\cite{blattmann2023stable} implements a two-stage pipeline where temporal upsampling through frame interpolation increases frame count while spatial super-resolution enhances image quality.  
Additionally, I2VGen-XL~\cite{zhang2023i2vgenxl} exemplifies the spatial cascade paradigm. 
The base stage generates semantically consistent videos, establishing the core content and motion. 
A refinement model then upscales to higher resolution, avoiding redundant feature learning. 
This design decouples semantic understanding   from visual quality, allowing  independent optimization. 
As models grow  sophisticated, recent works notice that spatial-temporal separation  creates boundaries. 
Therefore, MagicVideo-V2~\cite{wang2024magicvideo} introduces joint image-video training within a multi-stage framework that processes both modalities.   
This enables the model to leverage image data for improving spatial quality while learning temporal dynamics from videos. 
From another perspective, Lumiere~\cite{bartal2024lumiere} proposes the \textit{Space-Time U-Net} (ST U-Net), which applies downsampling and upsampling across spatial and temporal dimensions, compressing both time and space. 

%% file: sections/sec7_1_human_animation.tex
Applications of I2V models refer to tasks that use I2V-generated videos  to support a target pipeline, which is crucial for understanding the real impact of I2V.
\vspace{-0.5em}
\subsection{Character and Portrait Animation}
\label{sec:human_animation}
Character and portrait animation~\cite{xue2024follow,yang2024showmaker,wang2025unianimatedit,tan2025animatex,tu2025stableanimator,xu2025hunyuanportrait,wang2025animate,ma2024follow} represent one of the most active applications of I2V models. 
Specifically, MegActor-Sigma~\cite{yang2025megactor}, SkyReels-A1~\cite{qiu2025skyreels}, Hallo3~\cite{cui2025hallo3}, and DreamActor-M1~\cite{luo2025dreamactor} focus on portrait animation from a single reference image. 
Meanwhile, 
MimicMotion~\cite{zhang2025mimicmotion} targets human motion animation from a single person image. 
It produces motion-consistent videos under pose-based control, which supports applications such as dance, sports, and gesture animation.
Additionally, 
MIMO~\cite{men2024mimo} and Structural Video Diffusion~\cite{wang2025multi} broaden character video synthesis beyond single-subject animation, enabling multi-character storytelling and compositional content creation. 
Moreover, 
PersonalVideo~\cite{li2024personalvideo} targets subject-specific video customization, enabling repeated generation for the same identity while maintaining high  fidelity. 
In summary, these works position I2V as a practical animation engine that turns static character assets into editable, identity-consistent portrait and character videos.

\begin{figure*}[t]
    \centering
    \includegraphics[width=0.95\linewidth]{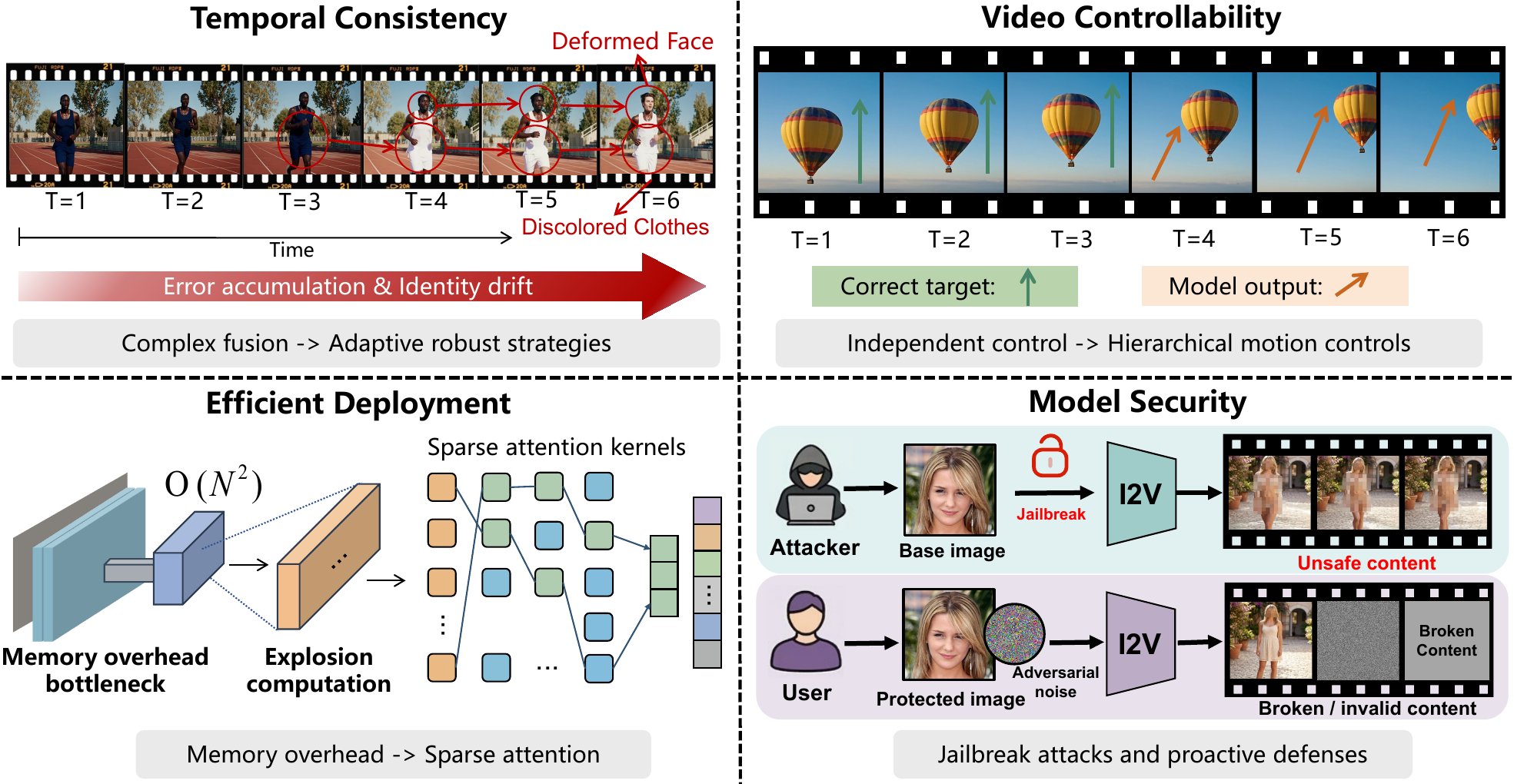}
    \caption{Four open problems and future research directions in diffusion I2V. }
    \label{fig:future}
\end{figure*}

 \vspace{-0.5em}
\subsection{Viewpoint Control}
\label{sec:camera_control}

Precise control over camera motion transforms I2V generation into active scene exploration, enabling view synthesis applications.   
To encode camera poses while ensuring 3D geometric consistency,  
CamCo~\cite{xu2024camco}  introduces Plucker coordinates as a dense, pixel-wise camera representation that encodes both intrinsic and extrinsic parameters. 
Building on this, CamI2V~\cite{zheng2024cami2v}  constrains cross-frame attention to epipolar lines, reducing model uncertainty under high noise levels. 
However, both CamCo~\cite{xu2024camco} and CamI2V~\cite{zheng2024cami2v} assume users can provide precise camera parameters, which is impractical. 
To address this, 
RealCam-I2V~\cite{li2025realcam}   shifts from relative-scale to absolute-scale  control through monocular metric depth estimation, enabling camera parameter specification in meaningful physical units.  MotionCanvas~\cite{xing2025motioncanvas} presents a   cinematic shot design that controls both camera viewpoint and object movement. 
In addition, certain works explore  efficiency improvements and expand  camera-controlled applications.  
SRENDER~\cite{chen2026efficient} generates only sparse keyframes, reconstructing the 3D scene representation using Gaussian splatting, and rendering intermediate frames through 3D graphics. 
Beyond this,  ViewCrafter~\cite{yu2024viewcrafter} generates high-fidelity novel views from single image by conditioning on point cloud renders.  
For 4D synthesis, 4DiM~\cite{watson2024controlling} introduces metric-scale camera pose control through a calibration pipeline. 
Building on these,  
DimensionX~\cite{sun2024dimensionx} decomposes video generation into spatial and temporal spaces,  Diffusion4D~\cite{liang2024diffusion4d} introduces a 3D-to-4D motion magnitude metric that measures dynamic strength and enables fine-grained control over motion intensity. 
DaS~\cite{gu2025diffusion} leverages 3D tracking videos as control signals, providing explicit temporal linkage superior to depth maps. 
RCM~\cite{wang2026rcm} tackles 3D character generation through a progressive training strategy that separately teaches pose canonicalization, viewpoint initialization, and character rotation.

\vspace{-0.5em}
\subsection{Video Media Production}
\label{sec:media_production}

The I2V model can be applied in video media production, where it transforms images into dynamic video content. 
By enabling personalized video creation and   scene transitions, it enhances the production of tailored videos. 
\textit{(1) Transition Generation.} 
Video media creators commonly need smooth transitions between visual concepts for predicting motion continuations. 
Therefore, 
TVG~\cite{zhang2024tvg} employs Gaussian process regression to model latent representations during transitions, providing a probabilistic framework for interpolation. 
Building on this, VTG~\cite{yang2025vtg} presents a unified framework handling four distinct transition tasks to correlate frame-wise latents. 
For short-to-long video generation, SEINE~\cite{chen2024seine} generates transition frames using random mask strategies, enabling coherent extension of video sequences. 
These capabilities enable smooth transitions in video storytelling for media production. 
\textit{(2) Personalized Generation. }
Creating videos featuring specific subjects with consistent appearance remains a challenge for video media production. 
Early methods adapt image customization techniques like \textit{Textual Inversion} (TI)~\cite{gal2022textual} and DreamBooth~\cite{ruiz2023dreambooth} to video generation, but often degrade motion dynamics. 
Hence, 
PIA~\cite{zhang2024pia} introduces plug-and-play animation modules that decouple appearance alignment from motion alignment, enabling effective  motion control while preserving personalized styles through inter-frame affinity hints. 
Meanwhile, 
CustomCrafter~\cite{wu2025customcrafter} addresses identity preservation through dynamic weighted video sampling, reducing LoRA weight during initial denoising steps and restoring it in final steps to maintain motion fluency.  PersonalVideo~\cite{li2024personalvideo} further advances the field through non-reconstructive training that directly supervises generated videos, with the isolated identity adapter injecting identity features  into spatial self-attention layers.  Besides, VideoBooth~\cite{jiang2024videobooth} enables tuning-free customization through image prompts using a coarse-to-fine embedding strategy, where coarse CLIP features provide semantic encoding and fine attention injection preserves spatial details through multi-scale latent incorporation. 
These approaches make it easier for I2V models to generate personalized content, thereby contributing to video media production.

%% file: sections/sec8_open_problems.tex
\section{Open Problems and Future Directions}
\label{sec:open_problems}

Despite remarkable progress in I2V diffusion models over the past years, numerous challenges remain unresolved. 
This section synthesizes the limitations of existing schemes, organizing them into future research directions, as seen in~\cref{fig:future}. 

\vspace{-0.5em}
\subsection{Temporal Consistency}
\label{sec:temporal_consistency}
Temporal consistency guarantees that transitions between frames are seamless, preserving the narrative structure and visual integrity of generated videos. 
While schemes like ConsistI2V~\cite{ren2024consisti2v} introduce first-frame attention and frequency noise initialization to reduce temporal drift, 
they still remain vulnerable to accumulated errors and lack temporal consistency.  
Sliding window methods like MagicInfinite~\cite{yi2025magicinfinite} and StableAvatar~\cite{tu2025stableavatar} require sophisticated fusion mechanisms for continuity, while autoregressive methods like MAGI-1~\cite{teng2025magi} and SkyReels-V2~\cite{chen2025skyreels} employ progressive noise scheduling and diffusion forcing. FramePack~\cite{zhang2025framepack} enables infinite-length generation through importance-aware compression, yet artifacts from aggressive downsampling remain visible in high-motion scenarios.
In summary, current methods still encounter challenges like error accumulation, complex fusion, and artifacts in high-motion scenarios. 
Future work will focus on developing more adaptive strategies to address these issues, potentially leveraging advanced learning techniques to prevent drift and improve continuity across longer sequences.

\vspace{-0.5em}
\subsection{Video Controllability}
\label{sec:controllability_interp}
Video controllability~\cite{yin2026focal} is essential for I2V models since it enables precise alignment between user intent and generated videos, ensuring predictable and editable outputs for downstream applications.
Trajectory control methods like Tora~\cite{zhang2025tora} and DragAnything~\cite{wu2024draganything} enable object path specification, but struggle when trajectories conflict with physical constraints. 
Meanwhile, 
camera control approaches like CamCo~\cite{xu2024camco} and RealCam-I2V~\cite{li2025realcam} achieve  geometric consistency through epipolar constraints, yet operate independently from object motion control, creating ambiguity when camera movement and object dynamics must be coordinated. 
The limitation behind these schemes is the inability to reason about hierarchical relationships between control modalities. 
Therefore, 
MotionCanvas~\cite{xing2025motioncanvas} makes initial progress by decomposing motion hierarchically into camera, global object, and local object components, 
but this decomposition was manually designed rather than learned.     
In summary, these advancements highlight the challenges of independently controlling video elements, and future solutions are supposed to focus on learning and integrating hierarchical motion controls for both objects and cameras to achieve true video controllability.

\vspace{-0.5em}
\subsection{Efficient Deployment}
\label{sec:efficiency_scale} 
Efficiency for I2V is crucial for enabling scalable and practical video generation under real-world computational constraints. 
In response, acceleration schemes like SVG~\cite{xi2025sparse}, UniCP~\cite{sun2025unicp}, LTX-Video~\cite{hacohen2024ltx}, and CausVid~\cite{yin2025causvid} 
have achieved improved quality-efficiency via training-free sparse attention kernels, unified caching and pruning, global self-attention
with a compressed spatiotemporal latent space, 
and distillation, respectively. 
Moreover, controlling memory consumption is critical for deploying I2V models, as it governs whether long-context spatiotemporal attention can be executed within practical device limits.  
For instance, 
FramePack~\cite{zhang2025framepack} maintains constant memory consumption for videos through importance-aware compression, but downsampling artifacts remain problematic. 
The slice VAE of 
EasyAnimate~\cite{xu2024easyanimate} enables temporal compression at encoding, but retraining VAE components adds overhead. 
Therefore, improving I2V efficiency is an important and practical avenue for future.

\vspace{-0.5em}
\subsection{Model Security}
\label{sec:security}
With the rapid evolution of I2V diffusion models, their security risks have also increasingly drawn attention.  
Gui~\etal~\cite{gui2025i2vguard} are the first to study unauthorized misuse of diffusion-based I2V models and propose protecting images with adversarial perturbations~\cite{goodfellow2014explaining} to prevent malicious I2V generation. 
In terms of attacks, Wang~\etal~\cite{wang2025runawayevil} provide the first study of jailbreak attacks~\cite{niu2024jailbreaking} on I2V models, successfully inducing Wan~\cite{wan2025open}, DynamiCrafter~\cite{xing2024dynamicrafter}, and CogVideoX~\cite{yang2025cogvideox} to generate malicious video contents. 
Therefore, research on attacks and defenses in I2V models will likely emerge as an important trend, and ensuring that I2V outputs are trustworthy and compliant is crucial for real-world deployment.

%% file: sections/sec9_conclusion.tex
\section{Summary}
\label{sec:summary}
This paper reviews diffusion I2V generation as a distinct research field. 
We examine the area through task formulation, architecture design, training paradigms, datasets, and evaluation metrics, and introduce a two-level taxonomy to show how existing methods evolve across modeling choices. We further summarize four key designs: condition encoding, temporal modeling, noise prior design, and spatial-temporal upsampling. 
The paper also discusses practical applications of this field in content creation.  
Simultaneously, we reveal that diffusion I2V generation still faces  challenges in temporal consistency, controllability, efficient deployment, and model security. 
Future progress requires not only stronger backbones, but also more principled designs for condition integration, motion generation, inference efficiency, and robustness.